\documentclass[graybox]{svmult}

\usepackage{mathptmx}
\usepackage{helvet}
\usepackage{courier}
\usepackage{graphicx}
\usepackage{makeidx}
\usepackage{multicol}
\usepackage{xcolor}
\usepackage{amsmath}
\usepackage{multirow}

\begin{document}

%
%

\title*{Super-Resolution for Selfie Biometrics: Introduction and Application to Face and Iris}
\author{Fernando Alonso-Fernandez, Reuben A. Farrugia, Julian Fierrez, Josef Bigun}
\institute{Fernando Alonso-Fernandez, Josef Bigun \at School of Information Technology (ITE), Halmstad University, Box 823, 30118 Halmstad, Sweden, \email{feralo@hh.se, josef.bigun@hh.se}
\and Reuben A. Farrugia \at Department of Communications and Computer Engineering (CCE), University of Malta, Malta \email{reuben.farrugia@um.edu.mt}
\and Julian Fierrez \at Escuela Politecnica Superior, Universidad Autonoma de Madrid, 28049 Madrid, Spain \email{julian.fierrez@uam.es}}
%
%
\maketitle


The lack of resolution has a negative impact on the performance of image-based biometrics.
Many applications which are becoming
ubiquitous in mobile devices do not operate in a controlled environment,
and their performance significantly drops due to the lack of pixel resolution,
since it decreases the amount of information available for recognition \cite{[Jain11a]}.

While many generic super-resolution techniques have been studied
to restore low-resolution images for biometrics \cite{Nasrollahi2014,[Thapa16]},
the results obtained are not always as desired.
Those generic super-resolution methods are usually
aimed to enhance the visual appearance of the scene.
However, producing an overall visual enhancement of biometric images
does not necessarily correlate with a better recognition performance \cite{[Alonso12a],[Galbally14]}.
Such techniques are designed to
restore generic images and therefore do not exploit the specific structure
found in biometric images (e.g. iris or faces),
which causes the solution to be sub-optimal \cite{[Farrugia17]}.
For this reason, super-resolution techniques have to be adapted to cater for the
particularities of images from a specific biometric modality \cite{[Baker02]}.

In recent years, there has been an increased interest in the application of
super-resolution to different biometric modalities, such as face iris, gait
or fingerprint \cite{[Nguyen18a]}.
This chapter presents an overview of recent advances in
super-resolution reconstruction of face and iris images,
which are the two prevalent modalities in selfie biometrics.
We also provide experimental results using several state-of-the-art reconstruction algorithms,
demonstrating the benefits of using super-resolution
to improve the quality of face and iris images prior to classification.
In the reported experiments, we study the application of super-resolution
to face and iris images captured in the visible range,
using experimental setups that represent well the selfie biometrics scenario.
The chapter begins with a general introduction to image resolution,
including the usual mathematical formulation,
a brief taxonomy of super-resolution methods, and performance metrics.
We then focus on face biometrics, describing recent super-resolution methods
adapted for this biometrics including experimental results.
Another section on iris super-resolution follows with a parallel structure.
The chapter ends with a summary and an outlook of future trends.

\section{Image Super-Resolution}

The performance of biometric recognition systems and the quality perceived by the human visual system (HVS) is significantly affected by the resolution of the image.
These images can be up-scaled using classical interpolation schemes used in several commercial software such as bilinear and bicubic interpolation \cite{Gonzalez2006}.
These methods use kernels that assume that the image data is either spatially smooth or band-limited and usually reconstruct blurred images  \cite{Thevenaz2000}.
More sophisticated interpolation methods were proposed in \cite{Carrato1996,Su2004} that manage to restore sharper images at the expense of generating visual artefacts in texture-less regions of the image.
While more advanced interpolation schemes manage to restore sharper images, they fail to reliably restore texture detail which  is important for biometric recognition systems.

Several researchers have  proposed more advanced techniques that recover the lost high-frequency information.
These methods usually formulate the problem using the following acquisition model

\begin{equation}
\mathbf{X} = \mathbf{D}\mathbf{B}\mathbf{Y} + \mathbf{\eta}
\label{eq:acquisitionmodel}
\end{equation}

\noindent where $\mathbf{X}$ is the observed low-resolution image, $\mathbf{B}$ is the blurring kernel, $\mathbf{D}$ is the down-sampling matrix, $\mathbf{\eta}$ represents additive noise and $\mathbf{Y}$ is the unknown high-resolution image to be estimated.

Existing super-resolution methods can be categorized in two groups: \romannumeral 1) \textit{Reconstruction}-\textit{based} super-resolution techniques,
which exploit the redundancies present in images and videos to estimate and restore an image,
and \romannumeral 2) \textit{Learning-based} super-resolution methods,
which treat the problem as an inverse problem and learn a mapping relation between the low- and high-resolution images.
More detail about each category is provided in the following subsections, while a comprehensive survey can be found in \cite{Nasrollahi2014}.

\subsection{Reconstruction-based methods}
\label{sec:reconstruction_methods}

Reconstruction-based algorithms try to address the aliasing artifacts that are present in the observed low-resolution images due to the under-sampling process.
Iterative back projection (IBP) methods \cite{Irani1990,Irani1993} use the acquisition model defined in Equation (\ref{eq:acquisitionmodel}).
These methods first register a sequence of low-resolution images over the high-resolution grid which are then averaged to estimate the high-resolution image.
IBP is then used to refine that initial solution.
To facilitate convergence and increase robustness against outliers,
the authors in \cite{Zomet2000,Farsiu2003} regularize the objective function using either smoothness or sparse constraints.

These methods were later on extended by considering different fusion kernels and including a de-blurring filter as a post-process e.g. using the Wiener Filter as suggested in \cite{Gonzalez2006}.
The authors in \cite{Farsiu2003b,Farsiu2004} have shown that the median fusion of the registered low-resolution images is equivalent to the maximum-likelihood estimation and results in a robust super-resolution algorithm if the motion between the low-resolution frames is translational.
Different data fusion techniques \cite{[Fierrez18]}
based on adaptive averaging \cite{Pham2006}, Adaboost classification \cite{Simonyan2008} and SVD-based filters \cite{Nasir2011} were also considered.
Probabilistic-based super-resolution techniques based on the Maximum-Likelihood (ML) \cite{Cheeseman1996, Schultz1996} and Maximum a-Posteriori (MAP) \cite{Hardie1997}
were proposed to estimate the high-resolution frame.

More recently, a framework that extends reconstruction-based super-resolution methods for the single image super-resolution problem was proposed in  \cite{Glasner2009}.
This method is based on the observation that patches in a natural image tend to  reoccur many times inside an image, both within the same scale as well as across different scales.

On the other hand,
Lin \textit{et. al}. \cite{Lin2004} have derived the theoretical limits of recon- struction-based super-resolution methods and proved that they can only achieve low magnification factors ($\leq 2$).
Moreover, these methods (except for the work in \cite{Glasner2009}) are only applicable for video sequences, i.e. they require several low-resolution
images as input, and in general they fail to restore dynamic non-rigid objects such as faces.

Due to the limitations of reconstruction-based super-resolution,
the research community is increasingly paying more attention to
learning-based super-resolution methods, which can recover more texture detail and achieve higher magnification factors.
They also have the advantage of only needing one image as input.

\subsection{Learning-based methods}
\label{sec:learning_methods}

\begin{figure*}[t]
\centering
\includegraphics[width=11cm]{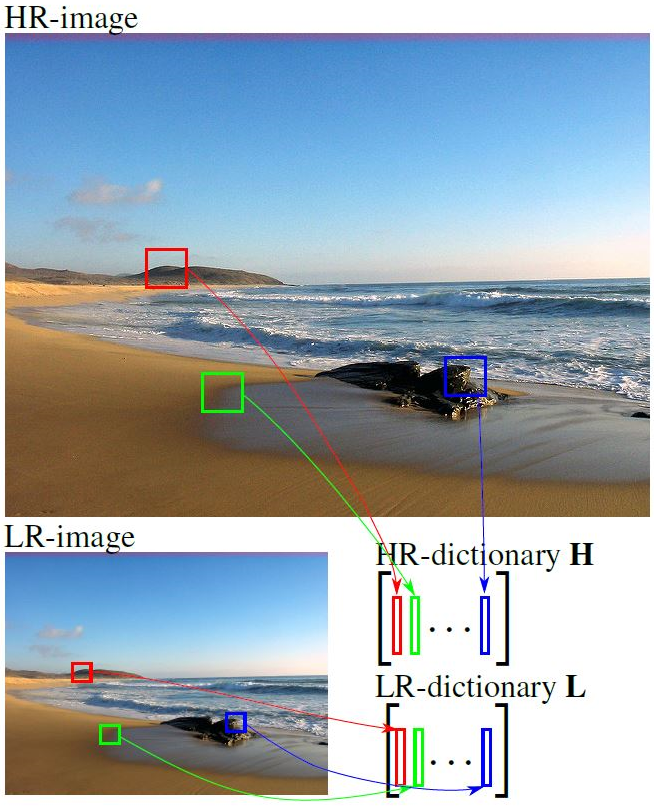}
\caption{Dictionary Construction for a learning-based super-resolution algorithm.}
\label{fig:LR_HR_Dict}
\end{figure*}

The seminal work of Freeman \cite{Freeman2002} presented the first example-based (\textit{a.k.a}. learning-based) super-resolution algorithm.
This class of methods employs a couple dictionary of low- and corresponding high-resolution patches which are constructed by collecting collocated patches from a set of low- and high-resolution training images. Figure \ref{fig:LR_HR_Dict} illustrates the principle of how the low-resolution $\mathbf{L}$ and high-resolution $\mathbf{H}$
dictionaries are constructed.
The authors in \cite{Freeman2002} then proposed to subdivide the input image into low-resolution patches that are traversed in raster-scan order.
At each step, a low-frequency patch is selected by a nearest neighbour search from the low-resolution dictionary $\mathbf{L}$.
The high-resolution patch is then estimated using the collocated patch in the high-resolution dictionary $\mathbf{H}$.
Markov Random Fields are then used to enforce smoothness across neighbouring patches.
The reconstructed high-resolution patches are then stitched together to form the high-resolution image.

The authors in \cite{Chang2004} observed that small patches from low- and high-resolution images form manifolds with similar local geometry in two distinct spaces.
They then use Local Linear Embedding (LLE) to find the $k$-closest neighbours from $\mathbf{L}$ to the $i$-th low-resolution patch $\mathbf{x}_i$ to form the sub-dictionary $\mathbf{L}_k$.
The reconstruction weights $\mathbf{w}$ are then computed using the following optimization problem

\begin{equation}
\mathbf{w} = \arg \min_ {\mathbf{w}}{||\mathbf{x}_i - \mathbf{L}_k \mathbf{w}||_2^2} \text{ subject to } \sum_j{w_j} = 1
\label{eq:LLE}
\end{equation}

\noindent which has a closed form solution. The high-resolution patch $\tilde{\mathbf{y}}_i$ is then reconstructed using

\begin{equation}
\tilde{\mathbf{y}}_i = \mathbf{H}_k \mathbf{w}
\end{equation}

\noindent where $\mathbf{H}_k$ correspond to the $k$ column vectors from $\mathbf{H}$ that correspond to the $k$-closest neighbours on the low-resolution manifold.
Several researchers have proposed different ways of estimating the combination weights $\mathbf{w}$, the most notorious one is to pose a sparsity constraint on the weights as done in \cite{Yang2008}

\begin{equation}
\mathbf{w} = \arg \min_ {\mathbf{w}}{||\mathbf{x}_i - \mathbf{L} \mathbf{w}||_2^2} \text{ subject to } ||\mathbf{w}||_1
\label{eq:SC}
\end{equation}

\noindent that can be computed in polynomial time using sparse coding solvers and is capable to outperform the neighbour-embedding scheme \cite{Chang2004}.
Later on, the same group has shown in \cite{Yang2010} that performance can be further improved using dictionary learning techniques that jointly train the low- and high-resolution dictionaries  to generate a more compact representation of the patch pairs which simply sample a large amount of image patch pairs as shown in Figure \ref{fig:LR_HR_Dict}.

The authors in \cite{Dong2013} have shown that sparse representations are affected by the distortions present in the low-resolution image, and are therefore not accurate enough to faithfully reconstruct the original image.
They then reformulate the sparse coding problem in (\ref{eq:SC}) as

\begin{equation}
\mathbf{w} = \arg \min_ {\mathbf{w}}{||\mathbf{x}_i - \mathbf{L} \mathbf{w}||_2^2} \text{ subject to } ||\mathbf{w}||_1 \text{ and } \sum_j{(w_j - \beta_j)^2}  \leq \epsilon
\label{eq:NCSR}
\end{equation}

\noindent where $\mathbf{\beta}$ is estimated from the sparse coding coefficients of neighbouring patches.

\begin{figure}[!h]
\centering
\begin{tabular}{cc}
\centering
\includegraphics[width=5.5cm]{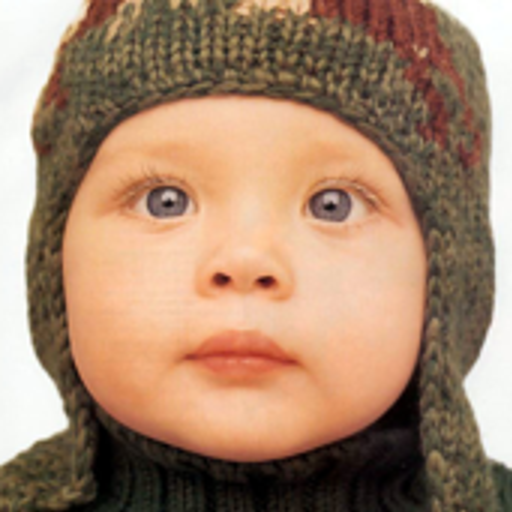} &
\includegraphics[width=5.5cm]{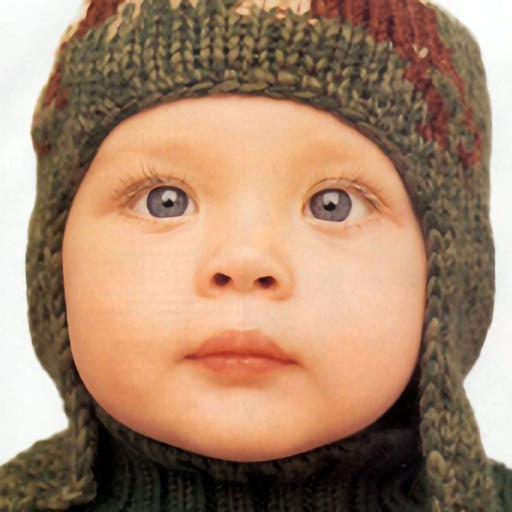}\\
\\
\footnotesize{Bicubic} & \footnotesize{VDSR}\\
\end{tabular}
\caption{Comparing the performance of a very deep CNN (VDSR) against bicubic interpolation.}
\label{fig:subjective}
\end{figure}

Deep convolutional neural networks (DCNN) were investigated recently for the generic super-resolution task. In \cite{Dong2015}, the authors present a shallow network consisting of just three convolutional layers, providing substantial improvement over sparse coding-based super-resolution methods.
This model poses the super-resolution problem as a regression problem and uses a DCNN to model a function $f(\mathbf{X: \theta})$ that minimizes the following loss function

\begin{equation}
L(\theta) = \sum_j{f((\mathbf{X}_j: \theta) - \mathbf{Y}_j})^2
\end{equation}

\noindent where $\mathbf{X}_j$ and $\mathbf{Y}_j$ represent a set of low- and corresponding high-resolution training images, $j$ is an index and $\theta$ are the hyperparameters of the network.
More recently, very deep architectures were proposed in \cite{Kim2016, Lim2017} which employ deeper architectures and residual learning, and are reported to provide state of the art performance.
The results in Figure \ref{fig:subjective} show the performance of VDSR \cite{Kim2016} against bicubic interpolation where it can be clearly seen that VDSR is able to recover sharper images.

\subsection{Performance Metrics}

To evaluate the performance of super-resolution methods,
the Peak Signal-to-Noise ratio (PSNR)
and the Structural Similarity (SSIM) index
between the enhanced and the corresponding high-resolution reference images
are usually employed \cite{[Thapa16]}.

The PSNR is a measure of the ratio between the maximum possible power of a signal
and the power of corrupting noise that affects the fidelity of its representation.
The signal in this case is the reference high-resolution image $\mathbf{Y}$,
and the noise is the error introduced in its estimation $\tilde{\mathbf{Y}}$ by the reconstruction algorithm.
Considering gray-scale images of size $N \times M$ and gray-values in the [0,255] range,
it is defined (in dBs) as:

\begin{equation}
\textrm{PSNR} (\mathbf{Y},\tilde{\mathbf{Y}}) = 10\log _{10} \left[ {\frac{255^2}{{\textrm{MSE}(\mathbf{Y},\tilde{\mathbf{Y}}) }}} \right]
\end{equation}

\noindent where $\textrm{MSE} (\mathbf{Y},\tilde{\mathbf{Y}})$ is the mean squared error given by

\begin{equation}
\textrm{MSE} (\mathbf{Y},\tilde{\mathbf{Y}}) = \left[ {{{\frac{1}{{NM}}\sum\limits_{i = 1}^M {\sum\limits_{j = 1}^N {\left| {\mathbf{Y} \left( {i,j} \right) - \tilde{\mathbf{Y}} \left( {i,j} \right)} \right|^2 } } }}} \right]
\end{equation}

A higher PSNR generally indicates that the reconstruction is of higher quality.
If the two images are identical, $\textrm{MSE} (\mathbf{Y},\tilde{\mathbf{Y}})=0$, in whose case $\textrm{PSNR} (\mathbf{Y},\tilde{\mathbf{Y}})=\infty$.
The PSNR is an estimation of the absolute error between two images.
The SSIM index, on the other hand, is a perception-based model that considers image degradation as a perceived change in structural information.
This is achieved by using first and second order statistics of gray values on local image windows.
It is computed on various windows of an image.
Given two collocated image windows $\mathbf{y}$ and $\tilde{\mathbf{y}}$, the SSIM index is computed as

\begin{equation}
\textrm{SSIM}\left({\mathbf{y},\tilde{\mathbf{y}}}\right) = \frac{{\left( {2\mu _\mathbf{y} \mu _{\tilde{\mathbf{y}}}  + c_1 } \right)\left( {2\sigma _{\mathbf{y}{\tilde{\mathbf{y}}}}  + c_2 } \right)}}{{\left( {\mu _\mathbf{y}^2  + \mu _{\tilde{\mathbf{y}}}^2  + c_1 } \right)\left( {\sigma _\mathbf{y}^2  + \sigma _{\tilde{\mathbf{y}}}^2  + c_2 } \right)}}
\end{equation}

\noindent where the parameter $\mu _\mathbf{y}$ ($\mu _{\tilde{\mathbf{y}}}$) is the average gray value of $\mathbf{y}$ ($\tilde{\mathbf{y}}$),
the parameter $\sigma _\mathbf{y}$ ($\sigma _{\tilde{\mathbf{y}}}$) is the variance of the gray values of $\mathbf{y}$ ($\tilde{\mathbf{y}}$),
and $\sigma _{\mathbf{y}{\tilde{\mathbf{y}}}}$ is the covariance of $\mathbf{y}$ and $\tilde{\mathbf{y}}$.
By default, $c_1=(0.01*255)^2$ and $c_2=(0.03*255)^2$ \cite{[Wang04]}.
Also, the window size is of 11$\times$11, which is weighted with a circular Gaussian filter of standard deviation 1.5
before calculating local statistics.
The SSIM index is computed for all pixels of the image, which are then averaged to obtain the SSIM index of the overall image.
The SSIM index is a decimal value between -1 and 1, and value 1 is only reachable in the case of two identical images.

The use of super-resolution techniques in general applications is aimed at improving the
overall visual perception and appearance. In biometrics, however, the aim is to improve
the recognition performance \cite{[Nguyen18a]}.
While PSNR or SSIM are
the standard metrics widely used in the super-resolution literature,
they are not necessarily good predictors of the
recognition accuracy.
Human and machine evaluations of image quality may differ,
and human judgement may not be relevant to biometric algorithms \cite{[Alonso12a]}.
For this reason, reporting recognition performance using enhanced
images is necessary to evaluate the goodness of the reconstruction
algorithm in a biometric context.

\section{Face Super-Resolution}
\label{sec:face_hallucination}

In their seminal work in 2000, Baker and Kanade \cite{Baker2000} exploited the fact that human face images are a relatively small subset of natural scenes and introduced the concept of class-based super-resolution \textit{i.e}. only facial images are used to learn the coupled dictionaries $\mathbf{L}$ and $\mathbf{H}$.
This method employs a pyramid-based algorithm to learn a prior on the derivative of the high resolution facial images as a function of the spatial location in the image and the information from higher levels of the pyramid and the solution is derived using the MAP algorithm.
The authors in \cite{Hu2011} observed that similar faces share similar local structure and synthesize missing pixels using a linear combination of spatially neighbouring pixels.
This method was extended in \cite{Li2014} where they exploit the sparse nature of the pixel structure.
However, the performance of these reconstruction-based
methods significantly degrades when considering large magnification factors where the local pixels structure is degraded.

\subsection{Face Eigentransformation}
\label{subsec:face_eigen}

Face representation models were used in \cite{Wang2005,Park2008,Chakrabarti2007} to derive a set of low- and high-resolution prototypes.
The low-resolution face image is reconstructed using a weighted combination of low-resolution prototypes, and the learned weights are used to combine the high-resolution prototypes to synthesize the high-resolution face image.
To explain this principle we take the classical Eigentransformation method \cite{Wang2005} which was used as a baseline in several studies.
The low-resolution and high-resolution mean faces ($\mathbf{m}_L$ and $\mathbf{m}_H$ respectively) are computed as

\begin{equation}
\hat{\mathbf{m}}_L = \frac{1}{M}\sum_{i=1}^{M}{\mathbf{L}_i} \quad \text{ and } \quad \hat{\mathbf{m}}_H = \frac{1}{M}\sum_{i=1}^{M}{\mathbf{H}_i}
\label{eq:mean_face}
\end{equation}

\noindent where $M$ is the number of training faces and the notation $\mathbf{K}_i$ denotes the $i$-th column-vector of matrix $\mathbf{K}$.
The coupled dictionaries are then centred using

\begin{equation}
\bar{\mathbf{L}} = \mathbf{L} - \mathbf{m}_L \quad \text{ and } \bar{\mathbf{H}} = \mathbf{H} - \mathbf{m}_H
\label{eq:centering}
\end{equation}

Given an input low-resolution image $\mathbf{X}$,
it
can be approximated using a weighted combination of centred faces using

\begin{equation}
\tilde{\mathbf{X}} = \bar{\mathbf{L}} \mathbf{w} + \mathbf{m}_L
\label{eq:reconstruct_LRET}
\end{equation}

\noindent where

\begin{equation}
\mathbf{w} = \mathbf{V}_L \boldsymbol{\Lambda}_L^{-\frac{1}{2}} \mathbf{E}^T (\mathbf{X}-\mathbf{m}_L)
\label{eq:ET_weights}
\end{equation}

\noindent where $\mathbf{V}_L$ is the eigenvector matrix, $\boldsymbol{\Lambda}_L$ is the eigenvalue matrix and $\mathbf{E}$ is the eigenface matrix which are derived by computing PCA on the covariance matrix $\mathbf{C}_L = \bar{\mathbf{L}}^T\bar{\mathbf{L}}$.
The reconstruction of the high-resolution face image is done by simply replacing the low-resolution dictionary $\bar{\mathbf{L}}$ with the high-resolution dictionary $\bar{\mathbf{H}}$ and the low-resolution mean face $\mathbf{m}_L$ with the high-resolution mean face $\mathbf{m}_H$ in (\ref{eq:reconstruct_LRET}), which is therefore computed using

\begin{equation}
\tilde{\mathbf{Y}} = \bar{\mathbf{H}} \mathbf{w} + \mathbf{m}_H
\label{eq:synthesize_LRET}
\end{equation}

The above methods are able to hallucinate missing information by exploiting the global facial structure.
Nevertheless, the faces restored using these methods are generally noisy and their quality is usually inferior to bicubic interpolation.
This is mainly attributed to the fact that the dimension of the face image is much larger than the number of training examples which makes the dictionaries undercomplete.

\begin{figure*}
\centering
\includegraphics[width=11cm]{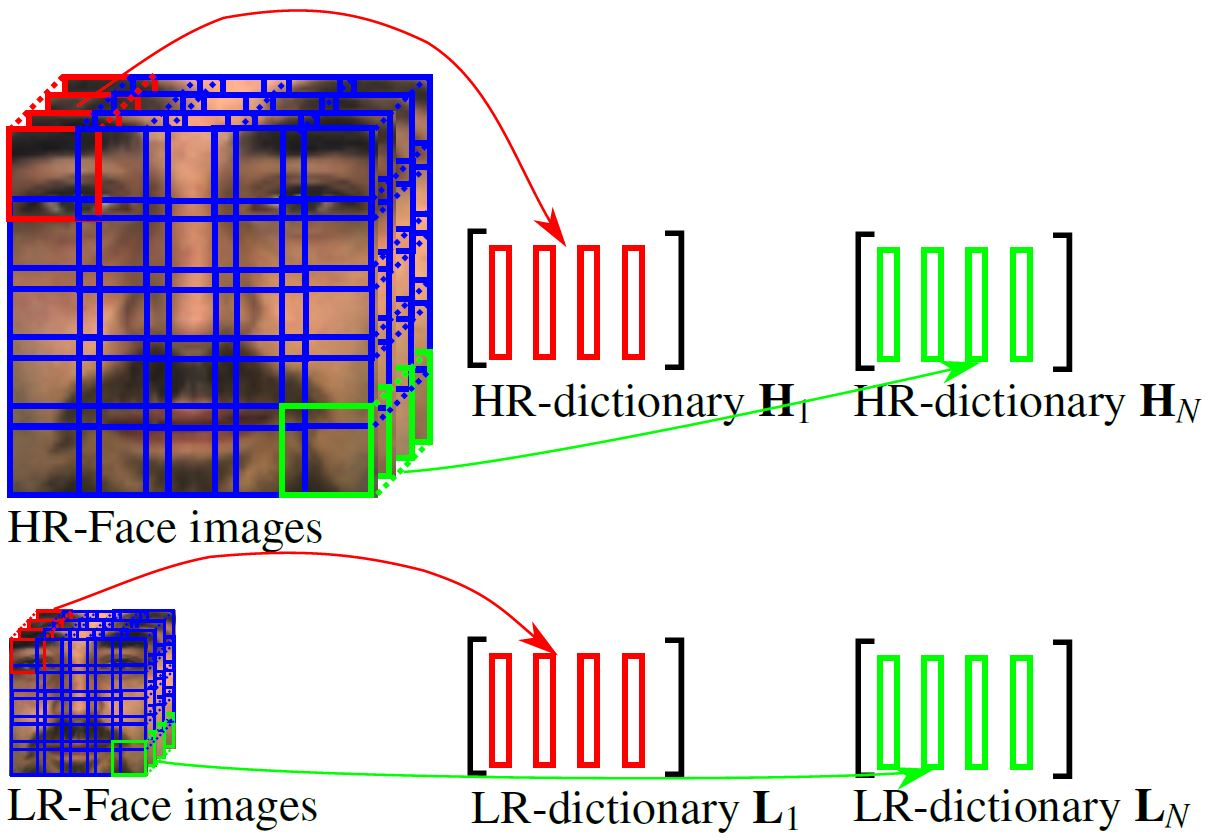}
\caption{Dictionary Construction using the Position-Patch principle.}
\label{fig:PositionPatch_Dict}
\end{figure*}

More recently, the authors in \cite{Ma2009} exploited the structure of the face and constructed position-dependent dictionaries, as show in Figure \ref{fig:PositionPatch_Dict}.
Face images are first aligned using affine transformation such that the eyes and mouth centres are aligned and then they are divided into overlapping patches.
Then they construct a coupled dictionary for each patch-position.
In the example in Figure \ref{fig:PositionPatch_Dict}, the high-resolution dictionary $\mathbf{H}_1$ (marked in red) consists of a vectorized representation of top-left position of all the $N$ high-resolution  images while the corresponding low-resolution dictionary $\mathbf{L}_1$ is composed of the vectorized representations of the $N$ low-resolution images.
This simple extension reduces the dimensionality of the problem and reduces the possibility of over fitting.
During testing, the low-resolution input image $\mathbf{X}$ is dissected into a set of overlapping patches that we shall denote as $\mathbf{x}_j$ where $j \in [1, N]$ represents the patch-position.
For each position-patch $j$, the authors use the coupled low- and high-resolution dictionaries $\mathbf{L}_j$ and $\mathbf{H}_j$ respectively.
Each patch is restored independently and then stitched together by averaging overlapping pixels.

The authors in \cite{Ma2009} proposed to formulate the restoration of a patch as a least squares problem

\begin{equation}
\mathbf{w}_j = \arg \min_ {\mathbf{w}_j} ||\mathbf{H}_j - \mathbf{L}_j \mathbf{w}_j ||_2^2
\label{eq:position_patch}
\end{equation}

\noindent which has the following closed form solution

\begin{equation}
\mathbf{w}_j = \left( \mathbf{L}_j^T\mathbf{L}_j \right)^\dagger\mathbf{L}_j^T \mathbf{H}_j
\end{equation}

\noindent where $\dagger$ stands for the pseudo-inverse operator.
Several researchers have extended the position-patch method using  different objective functions. The authors in \cite{Jung2011} have formulated the weight estimation problem using sparse coding

\begin{equation}
\mathbf{w}_j =\arg \min_ {\mathbf{w}_j} ||\mathbf{H}_j - \mathbf{L}_j \mathbf{w}_j ||_2^2 \quad \text{ subject to } \quad ||\mathbf{w}_j ||_1
\label{eq:sparse_position_patch}
\end{equation}

\noindent which enforces the combination weights $\mathbf{w}_j$ to be sparse.
This regularization allows deriving sharper facial images and is more robust to outliers.

\subsection{Local Iterative Neighbour Embedding}
\label{subsec:face_LINE}

One drawback of the methods of Section~\ref{subsec:face_eigen} is that they assume that low- and high-resolution manifolds have similar local geometrical structure.
Reconstruction weights are estimated on the low-resolution manifold, and they are simply transferred to the high-resolution manifold.
However, the low-resolution manifold is distorted by the one-to-many relationship between low- and high resolution patches \cite{Su2005,Li2009}.
Therefore, the reconstruction weights estimated on the low-resolution manifold do not necessarily correlate
with the actual weights needed to reconstruct the unknown high-resolution patch on the high-resolution manifold.

Motivated by this observation, the authors in \cite{Li2009,Huang2010} derive a pair of projection matrices that can be used to project both low- and high-resolution patches on a common coherent subspace.
However, the dimension of the coherent sub-space is equal to the lowest rank of the low- and high-resolution dictionary matrices.
Therefore, the projection from the coherent sub-space to the high-resolution manifold is ill-conditioned.

This ill-conditioning is overcome in the Locality-constrained Iterative Neighbour Embedding (LINE) method presented in \cite{Jiang2014} as follows.
They first estimate the high-resolution patch $\mathbf{v}_{0,0}$
by up-scaling the low-quality patch $\mathbf{x}_j$ using bicubic interpolation,
and initialize the intermediate dictionary as $\mathbf{L}_j^{\{0\}} = \mathbf{L}_j$.
This iterative method has an outer-loop indexed by $b \in [0, B-1]$ and an inner-loop indexed by $c \in [0,C-1]$.
For every iteration of the inner loop,
the supports $\mathbf{s}$ (\textit{i.e}. the column-vectors) of $\mathbf{H}_j$ are derived
as the $k$-nearest neighbours of $\mathbf{v}_{b,c}$. The combination weights are then computed using

\begin{equation}
\mathbf{w} = \arg \min_ \mathbf{w} ||\mathbf{x}_j-\mathbf{L}_j^{\{b\}}(\mathbf{s})\mathbf{w}||_2^2 + \tau ||\mathbf{d}(\mathbf{s}) \odot \mathbf{w}||^2_2
\label{eq:LINE}
\end{equation}

\noindent where $\tau$ is a regularization parameter, $\odot$ is the element-wise multiplication operator and $\mathbf{d(\mathbf{s})}$ measures
the Euclidean distance between the restored patch $\mathbf{v}_{b,c}$ and the $k$-nearest neighbours column-vectors from the high-resolution dictionary $\mathbf{H}_j$.
This optimization problem has an analytical solution and the high-resolution patch is updated using

\begin{equation}
\mathbf{v}_{b,c+1} = \mathbf{H}_j(\mathbf{s})\mathbf{w}
\end{equation}

Once all iterations of the inner-loop are completed, the intermediate dictionary $\mathbf{L}_j^{\{b+1\}}$ is updated using a leave-one-out methodology as described in \cite{Jiang2014} and the inner-loop is repeated. The final estimate of the high-resolution patch is then simply $\mathbf{v}_{B,C-1}$.

\subsection{Linear Model of Coupled Sparse Support}
\label{subsec:face_LMCSS}

While the method of Section~\ref{subsec:face_LINE} iteratively updates the low-resolution dictionary to restore the geometrical structure in the low-resolution manifold, it cannot guarantee to converge to an optimal solution.
Farrugia \textit{et al}. \cite{Farrugia2017} later presented the Linear Model of Coupled Sparse Support (LM-CSS) which learns linear models based on the local geometrical structure on the high-resolution manifold rather than on the low-resolution manifold.
For this, in a first step, the low-resolution patch is used to derive a globally optimal estimate of the high-resolution patch.
This is equivalent to solving the following problem

\begin{equation}
\boldsymbol{\Phi} = \arg \min_ {\boldsymbol{\Phi}} ||\mathbf{H}_j - \boldsymbol{\Phi}\mathbf{L}_j||_2^2 \quad \text{ subject to } ||\boldsymbol{\Phi}||_2^2 \leq \epsilon
\label{eq:lmcss}
\end{equation}

\noindent which can be solved using Ridge Regression. This approximated solution is close in Euclidean space to the ground-truth but is generally smooth and lacks the texture details needed by state-of-the-art face recognizers.
The authors then search for the sparse support that best estimates the first approximated solution on the high-resolution manifold where the geometric structure of manifold is intact.
The derived support is then used to extract the atoms from the coupled low- and high-resolution dictionaries $\mathbf{L}_j$ and $\mathbf{H}_j$ that are most suitable to learn an up-scaling function for every position patch.
The second step reformulates the problem as in Equation (\ref{eq:lmcss}), where only a subset of the column vectors, defined by the support, are used to find the solution.
This work also demonstrates that sparsity leads to sharper solutions and generally results in higher recognition accuracies.
The same authors have also demonstrated that these super-resolution techniques can be applied to restore compressed low-resolution facial images \cite{Farrugia2016}.

\subsection{Results}
\label{subsec:face_results}

In the experiments reported here, we consider a composite dataset which includes images from both Color FERET and Multi-PIE datasets, where only frontal facial images were considered. One image per subject was randomly selected, resulting in a dictionary of 1203 facial images. The gallery for the evaluation included images from both FRGC-V2 (controlled environment) and MEDS datasets. One unique image per subject was randomly selected, providing a gallery of 889 facial images. The probe images were taken from the FRGC (uncontrolled environment), where two images per subject were included, resulting in 930 probe images. All the images were registered using affine transformation computed on landmark points of the eyes and mouth centers such that the distance between the eyes is set to 40 pixels. The probe and low-resolution dictionary images were down-sampled to the desired scale using MATLAB's \emph{imresize} function.

\begin{figure*}[!ht]
\begin{center}
\begin{tabular}{ccccccc}
\centering

\includegraphics[width=.95\textwidth]{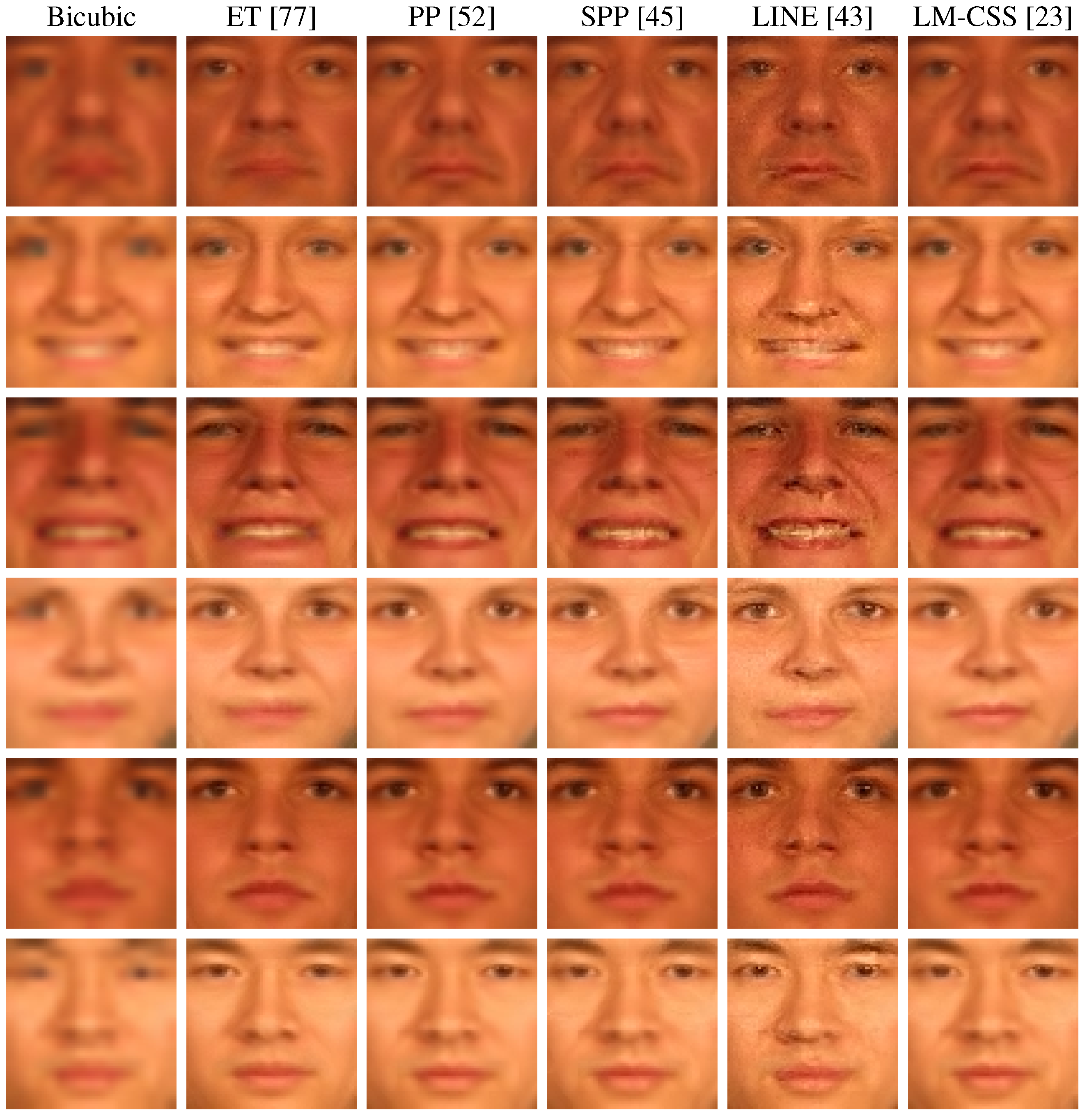}

\end{tabular}
\end{center}
\caption{Super-resolved face images using different face super-resolution techniques with a magnification factor of $\times 4$ with an inter-eye distance of 10 pixels.}
\label{fig:results-face}
\end{figure*}

The results in Table \ref{tbl:table_quality_face} and \ref{tbl:table_recognition_face} evaluate the performance of different face super-resolution techniques mentioned in this chapter in terms of both quality (PSNR and SSIM) and recognition performance (rank-1 and Area Under the Curve) respectively.
It can be seen that the global Eigentransformation method \cite{Wang2005} most of the time performs worse than bicubic.
This can be explained by the fact that while it reconstructs a face that is visually more pleasing than the ones obtained using bicubic interpolation, it fails to reliably recover the local texture detail (see example images in Figure \ref{fig:results-face}).
On the other hand, the remaining patch-based schemes outperform bicubic interpolation in terms of both quality and recognition performance.
The VGG-Face CNN face recognition system (DeepFaces) is also found to be particularly fragile, performing considerably worse than LBP for any resolution. For example, with a magnification factor of just 2 (Inter Eye distance = 20), its rank-1 accuracy is equal or below 40\% for any reconstruction technique.
It can also be noticed that while Position-Patch \cite{Ma2009}  achieves higher PSNR and SSIM compared to Sparse Position Patch \cite{Jung2011}, LINE \cite{Jiang2014} and LM-CSS with $k = 50$ \cite{Farrugia2017}, it does not perform well in terms of face recognition performance.
The authors in \cite{Farrugia2017} show experimentally using different face recognizers that the PSNR and SSIM metrics do not correlate well with the recognition performance since they are biased to provide high scores to blurred images.
It can be seen in Figure \ref{fig:results-face} that the images restored via Position-Patches are more blurred, which harms the recognition performance.
They also showed that sparse-based methods \cite{Jung2011, Jiang2014, Farrugia2017} are able to better preserve the texture detail and thus are able to achieve higher recognition performance.
The results in Figure \ref{fig:results-face} also show that the LINE method generally reconstructs sharp facial images, although they tend to be noisy.
This noise does not seem to harm the recognition performance but it may make it hard for a human to recognize a person from such noisy images.

\begin{table*}[!tb]
\centering
\caption{Summary of the Quality Analysis results using the PSNR and SSIM metrics on the FRGC dataset.}
\label{tbl:table_quality_face}
\begin{tabular}{|l||c|c||c|c||c|c||c|c|}
\multicolumn{1}{c}{} & \multicolumn{8}{c}{\bf{Inter Eye distance}} \\ \cline{2-9}
\multicolumn{1}{c}{} & \multicolumn{2}{c}{\bf{8}} & \multicolumn{2}{c}{\bf{10}} & \multicolumn{2}{c}{\bf{15}} & \multicolumn{2}{c}{\bf{20}}\\
\cline{2-9}

\multicolumn{1}{c||}{\bf{SR Method}} & \bf{PSNR} & \bf{SSIM} & \bf{PSNR} & \bf{SSIM} & \bf{PSNR} & \bf{SSIM} & \bf{PSNR} & \multicolumn{1}{c|}{\bf{SSIM}} \\
\hline

Bicubic 							 & 24.0292 & 0.6224 & 26.2024 & 0.7338 & 25.2804 & 0.7094 & 28.6663 & 0.8531 \\
Eigentransformation \cite{Wang2005} & 24.3958 & 0.6496 & 26.8645 & 0.7504 & 24.9374 & 0.6724 & 27.7883 & 0.7892 \\
Neighbour Embedding \cite{Chang2004}& 26.9987 & 0.7533 & 27.9560 & 0.7973 & 29.9892 & 0.8714 & 31.6301 & 0.9122 \\
Position-Patches \cite{Ma2009}       & 27.3044 & 0.7731 & 28.2906 & 0.8145 & 30.1887 & 0.8785 & 31.7192 & 0.9143 \\
Sparse Position-Patches \cite{Jung2011}       & 27.2500 & 0.7666 & 28.2219 & 0.8100 & 30.1290 & 0.8767 & 31.7162 & 0.9146 \\
LINE \cite{Jiang2014}               & 27.0927 & 0.7591 & 28.0253 & 0.8009 & 30.0471 & 0.8727 & 31.6970 & 0.9131 \\
LM-CSS \cite{Farrugia2017} ($k = 50$) & 27.1307 & 0.7679 & 28.1078 & 0.8093 & 30.0240 & 0.8761 & 31.6875 & 0.9139
\\
LM-CSS \cite{Farrugia2017} ($k = 150$)                 & \bf{27.4866} & \bf{0.7802} & \bf{28.4200} & 0.8009 & \bf{30.3431} & \bf{0.8845} & \bf{31.9610} & \bf{0.9209} \\
\hline
\end{tabular}
\end{table*}

\begin{table*}[tb]
\centering
\caption{Summary of the Rank-1 recognition performance and Area Under the Curve (AUC) metric using the LBP \cite{Ahonen2006} and DeepFaces \cite{[Parkhi15]} face recognition algorithm on the FRGC dataset.}
\label{tbl:table_recognition_face}
\begin{tabular}{|l|l||c|c||c|c||c|c||c|c|}
\multicolumn{2}{c}{} & \multicolumn{8}{c}{\bf{Inter Eye distance}} \\ \cline{3-10}
\multicolumn{2}{c}{} & \multicolumn{2}{c|}{\bf{8}} & \multicolumn{2}{c|}{\bf{10}} & \multicolumn{2}{c|}{\bf{15}} & \multicolumn{2}{c}{\bf{20}}\\ \cline{3-10}

\multicolumn{1}{c|}{\bf{SR Method}} & \multicolumn{1}{c|}{\bf{Comparator}} & \bf{rank-1} & \bf{AUC} & \bf{rank-1} & \bf{AUC} & \bf{rank-1} & \bf{AUC} & \bf{rank-1} & \bf{AUC} \\

\hline

Bicubic & LBP & 0.3065 & 0.9380 & 0.5032 & 0.9598 & 0.6065 & 0.9708 & 0.7054 & 0.9792 \\ \cline{2-10}
& DeepFaces & 0.0000 & 0.5296 & 0.0000 & 0.5337 & 0.0258 & 0.6223 & 0.1903 & 0.7157 \\ \hline \hline

Eigentrans- & LBP & 0.2559 & 0.9390 & 0.4516 & 0.9554 & 0.5624 & 0.9633 & 0.6495 & 0.9688 \\ \cline{2-10}
formation \cite{Wang2005} & DeepFaces & 0.0151 & 0.5794 & 0.0194 & 0.5954 & 0.0398 & 0.6187 & 0.1226 & 0.6612 \\ \hline \hline

Neighbour  & LBP & 0.5548 & 0.9635 & 0.6398 & 0.9712 & 0.7215 & 0.9795 & 0.7559 & 0.9830 \\ \cline{2-10}
Embedding \cite{Chang2004} & DeepFaces & 0.0151 & 0.5940 & 0.0602 & 0.6290 & 0.2086 & 0.6970 & 0.4075 & 0.7562 \\ \hline \hline

Sparse Position & LBP & 0.5677 & 0.9649 & 0.6441 & 0.9721 & 0.7247 & 0.9803 & 0.7570 & 0.9830 \\ \cline{2-10}
Patches \cite{Jung2011} & DeepFaces & 0.0161 & 0.5870 & 0.0419 & 0.6198 & 0.1624 & 0.6880 & 0.3796 & 0.7553 \\ \hline \hline

Position &  LBP & 0.4699 & 0.9588 & 0.5849 & 0.9675 & 0.6849 & 0.9782 & 0.7312 & 0.9812 \\ \cline{2-10}
Patches \cite{Ma2009} & DeepFaces & 0.0161 & 0.5914 & 0.0398 & 0.6201 & 0.1785 & 0.6878 & 0.3559 & 0.7408 \\ \hline \hline

LINE \cite{Jiang2014} & LBP & 0.5925 & 0.9647 & 0.6559 & 0.9714 & 0.7323 & \textbf{0.9804} & 0.7677 & \textbf{0.9833} \\ \cline{2-10}
& DeepFaces & 0.0312 & 0.6036 & 0.07 10 & 0.6385 & 0.2161 & 0.7050 & 0.4172 & 0.7630 \\ \hline \hline

LM-CSS \cite{Farrugia2017} & LBP & \textbf{0.6032} & \textbf{0.9658} & \textbf{0.6581} & \textbf{0.9722} & \textbf{0.7398} & 0.9798 & \textbf{0.7742} & \textbf{0.9833} \\ \cline{2-10}
($k = 50$) & DeepFaces & 0.0172 & 0.5874 & 0.0484 & 0.6293 & 0.1914 & 0.7015 & 0.4022 & 0.7609 \\ \hline \hline

LM-CSS \cite{Farrugia2017} & LBP  & 0.5452 & 0.9644 & 0.6344 & 0.9710 & \textbf{0.7398} & 0.9801 & 0.7602 & 0.9831 \\ \cline{2-10}
($k = 150$) & DeepFaces & 0.0151 & 0.5890 & 0.0527 & 0.6291 & 0.2108 & 0.7011 & 0.3828 & 0.7578 \\ \hline \hline

\hline
\end{tabular}
\end{table*}

One of the major problems in these methods is that they assume that the face images are aligned and cannot be applied directly to restore faces with random pose and orientation.
The authors in \cite{Farrugia2017} presented a simple method that registers the faces in the dictionary where a set of landmark points are used to register the dataset with the low-resolution image using piecewise affine transformation. Any face super-resolution method described in this chapter can then be used to restore faces with unconstrained poses.
However, this approach is computationally intensive and it is difficult for a user to accurately mark the landmark points on  very low-resolution images.
Following the success of deep-learning for generic super resolution,
the authors in \cite{Cao2017} applied deep learning to directly restore facial images
at arbitrary poses without the need for pre-registration.
The main advantage of this method is that it is very fast to compute, it does not need human interaction and it is able to restore the whole head including the hair region.
Nevertheless, while the results presented in the paper are promising, they were not assessed in terms of face recognition performance.

\section{Iris Super-Resolution}

Super-resolution was introduced to the iris modality in 2003 by Huang \emph{et al}. \cite{[Huang03]}.
This method learns the probabilistic relation between different frequency bands, which is
used to predict the missing high-frequency information of low-resolution images.
%
%
%
Reconstruction-based methods for iris started in 2006 with the work of
Barnard \emph{et al}. \cite{[Barnard06]}, where
they employed a multi-lens imaging hardware system to capture multiple iris images.
Reconstruction was done by modelling the least square inverse problem associated with Equation~(\ref{eq:acquisitionmodel}).
Later, Fahmy \cite{[Fahmy07]} proposed to estimate high-resolution images
using an auto-regressive model that fuses a sequence of low-resolution iris images.
While these two works super-resolve the original iris image, most of the
existing reconstruction-based methods employ the unwrapped polar image as input.
Several polar images are aligned and combined pixel-wise
to obtain a reconstructed image.
Given a set of $N$ polar iris images $\mathbf{X}_i$, the super-resolved image
$\tilde{\mathbf{Y}}$ is estimated as

\begin{equation}
\tilde{\mathbf{Y}}(x,y) = \frac{{\sum\limits_{i = 1}^N {w_i \mathbf{X}_i(x,y)} }}{{\sum\limits_{i = 1}^N {w_i } }}
\end{equation}

\noindent where $\tilde{\mathbf{Y}}(x,y)$ is the intensity value of at pixel $(x,y)$ of
the super-resolved image, $\mathbf{X}_i(x,y)$ is the intensity value at the same location
of the input image $i$, and ${w_i}$ are the combination weights.
The weights can be derived to simply compute the mean or median of the pixel values,
as in \cite{[Hollingsworth09a],[Nguyen10a],[Jillela11]}.
Other studies have proposed to incorporate quality measures
\cite{[Nguyen10],[Nguyen11],[Othman15],[Hsieh16]},
so more weight is given to frames with higher quality \cite{[Fierrez18a]}.
Recent reconstruction-based studies have proposed the use of
Gaussian Process Regression (GPR), Enhanced Iterated Back Projection (EIBP) \cite{[Deshpande17a]},
and Total Variation regularization algorithms \cite{[Deshpande17]}
to super-resolve polar frames.

Regarding learning-based methods, several algorithms have been proposed to learn
the mapping between low- and high-resolution images, for example
Multi-Layer Perceptrons \cite{[Shin09]},
Markov networks \cite{[Liu13]}, or
Bayesian modeling \cite{[Aljadaany15]}.
Some works have also proposed to super-resolve images in the feature space,
instead of the pixel domain.
This strategy has been followed with Eigeniris features \cite{[Nguyen11a]}
(similar to Eigenfaces proposed in \cite{[Turk91]})
and with the popular iris Gabor features \cite{[Nguyen12],[Nguyen13]}.
Recent studies also make use of Convolutional Neural Networks, such as
\cite{[Zhang16a],[Ribeiro17]}.

Despite the now extensive literature on iris super-resolution,
the majority of works have employed near-infrared data,
which is the prevalent illumination in commercial systems.
%
In the experiments reported in the present chapter,
we study the application of super-resolution to iris images
captured in the visible range using various smartphones,
using an experimental setup that represents well the selfie biometrics scenario.

\subsection{Iris Eigen-patches}
\label{subsec:iris_eigen}

\begin{figure}[t]
     \centering
     \includegraphics[width=.9\textwidth]{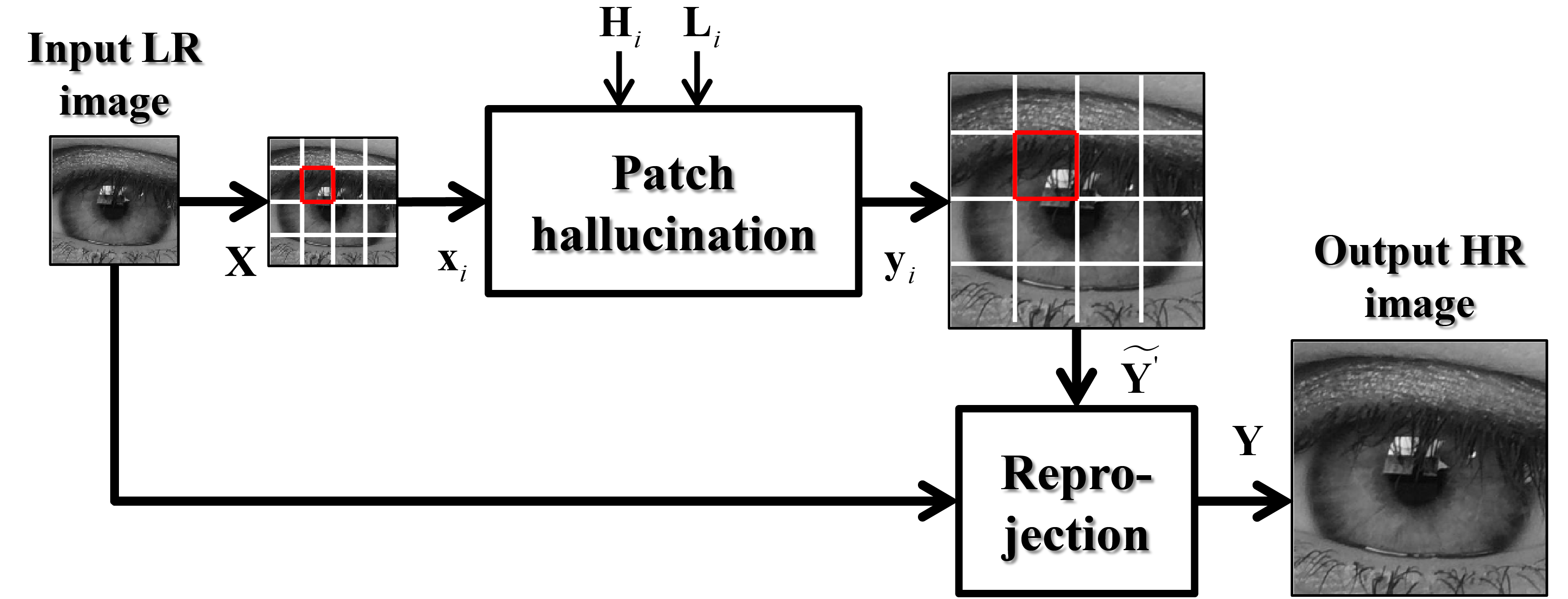}
     \caption{Block diagram of patch-based iris hallucination.}
     \label{fig:iris-patch-based}
\end{figure}

The work \cite{[Alonso15b]}
proposed the use of Principal Component Analysis (PCA)
Eigen-transformation of local image patches to compute a reconstructed iris image.
The technique is inspired by the system of \cite{[Chen14]} for face images.
It employs the Eigentransformation method
defined by Equations (\ref{eq:mean_face}-\ref{eq:synthesize_LRET}) \cite{Wang2005},
but applied to overlapping patches, as shown in Figure \ref{fig:iris-patch-based}.
The iris images are first resized via bicubic interpolation to have
the same iris radius, and then aligned by extracting a square
region around the pupil center.
Images are then divided into overlapping patches, and a coupled dictionary
is constructed for each patch-position.
Each patch is reconstructed separately, and a preliminary reconstructed image $\tilde{\mathbf{Y}'}$
is obtained by averaging the overlapping regions.
The authors in \cite{[Chen14]} also propose the incorporation of a
re-projection step to reduce artifacts and
make the output image more similar to the input image $\mathbf{X}$. The image
$\tilde{\mathbf{Y}'}$ is
re-projected to $\mathbf{X}$ via gradient descent using

\begin{equation}
\tilde{\mathbf{Y}}^{t + 1}  = \tilde{\mathbf{Y}}^t  - \tau \mathbf{U}\left( {\mathbf{B}\left(
{\mathbf{D}\mathbf{B}\tilde{\mathbf{Y}}^t  - \mathbf{X} } \right)} \right)
\end{equation}

\noindent where $\mathbf{U}$ is the up-sampling matrix.
The process stops when
$|\tilde{\mathbf{Y}}^{t + 1}-\tilde{\mathbf{Y}}^{t}|<\epsilon$.

\subsection{Local Iterative Neighbour Embedding}
\label{subsec:iris_LINE}

Recently, the work \cite{[Alonso17]} adapted the method described in Section \ref{subsec:face_LINE} to reconstruct iris images,
based on Multilayer Locality-constrained Iterative Neighbour Embedding (LINE)
of local image patches \cite{Jiang2014}.
In the mentioned work \cite{[Alonso17]}, and in the experiments reported here,
update of the intermediate dictionary has not been implemented.
On the other hand,
inspired by \cite{[Chen14]},
the re-projection step described in Section \ref{subsec:iris_eigen}
has been incorporated in the reconstruction of iris images
after the application of the LINE algorithm.


\subsection{Results}
\label{subsec:iris_results}

\begin{figure}[t]
     \centering
     \includegraphics[width=.9\textwidth]{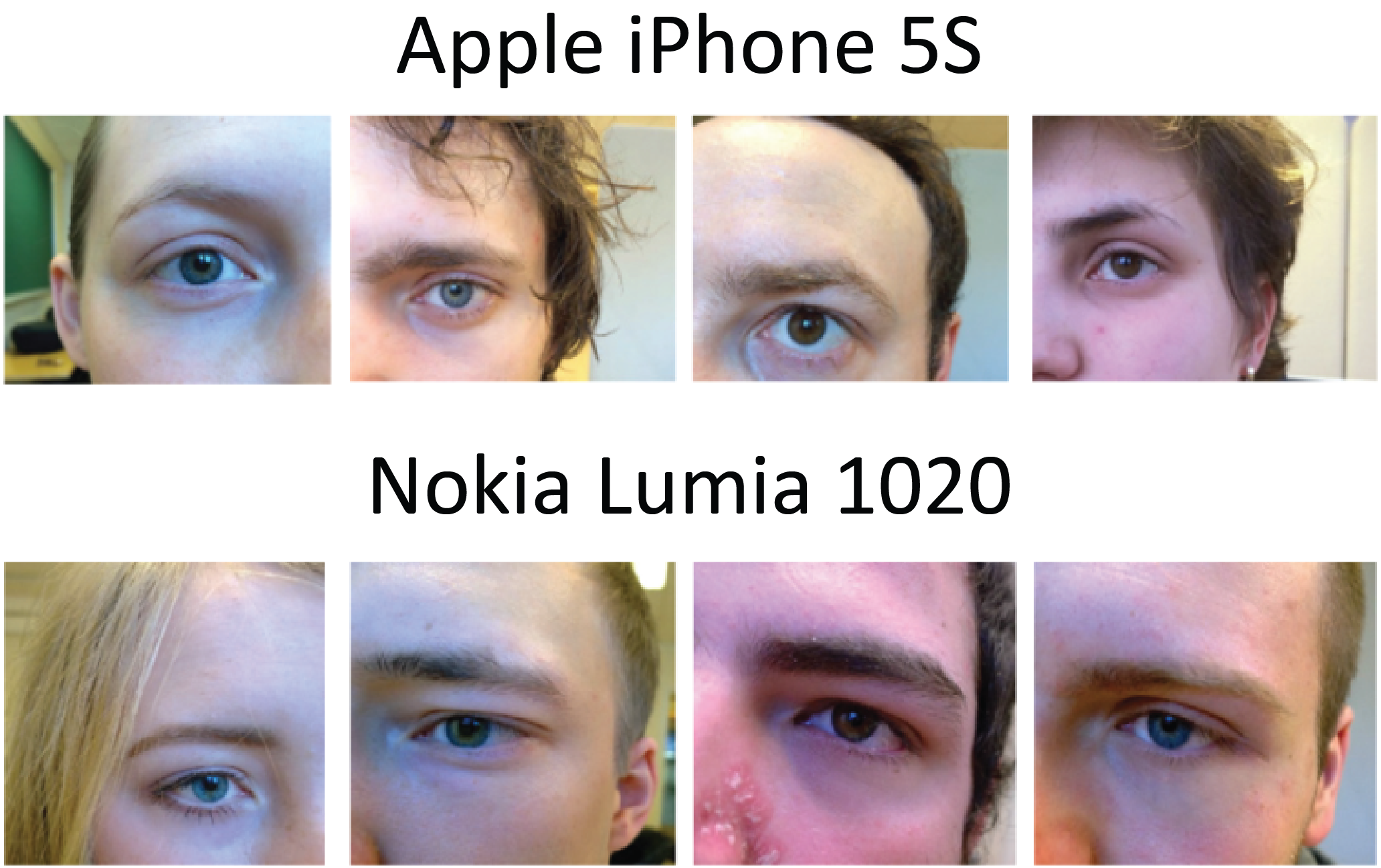}
     \caption{Sample images from VSSIRIS database \cite{[Raja14b]}.}
     \label{fig:db-iris}
\end{figure}

\begin{figure*}[t]
     \centering
     \includegraphics[width=.95\textwidth]{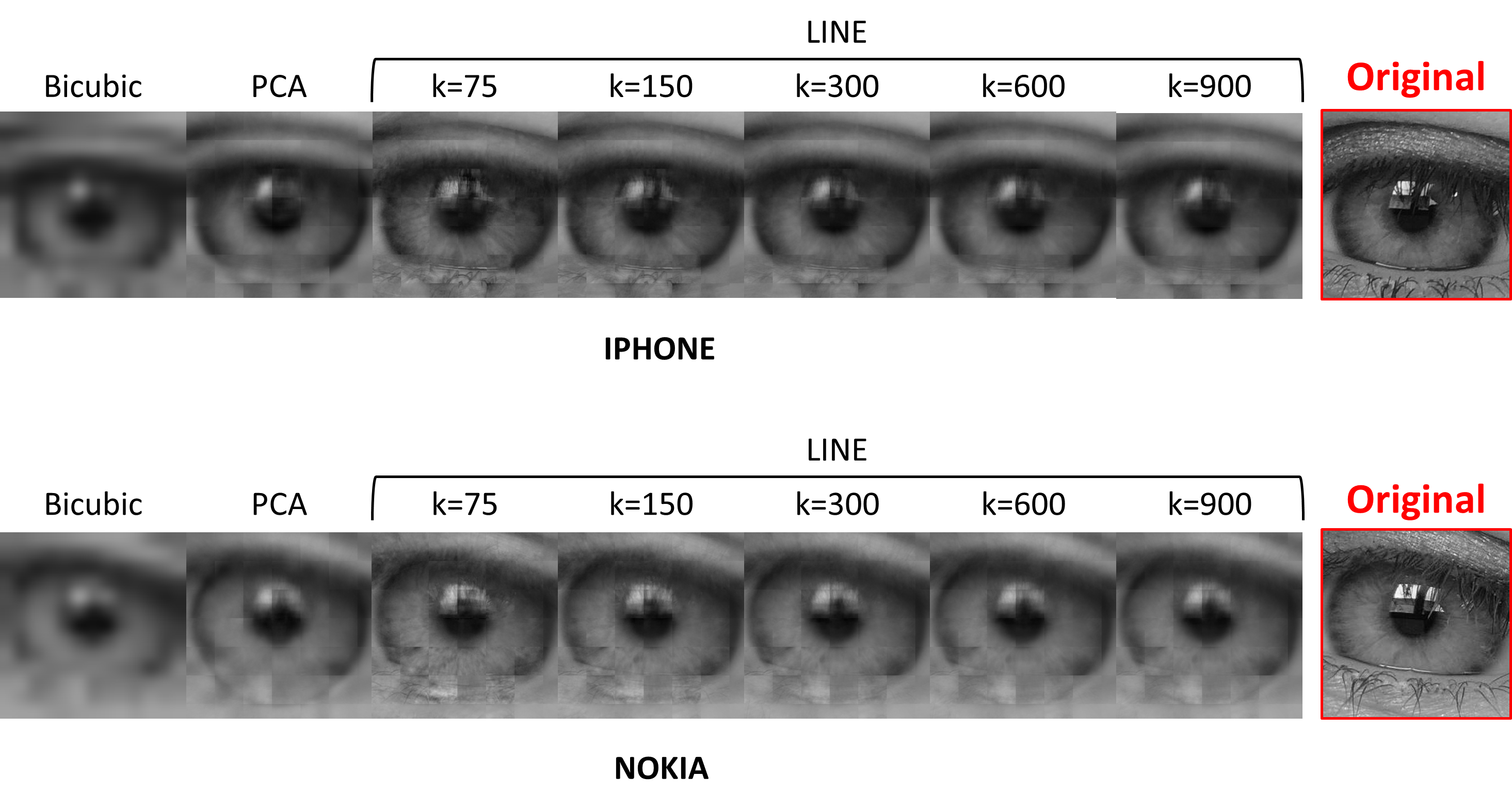}
     \caption{Super-resolved iris images using different iris super-resolution techniques with a magnification factor of $\times 22$.}
     \label{fig:images-example-iris}
\end{figure*}

We use the Visible Spectrum Smart-phone Iris (VSSIRIS) database
\cite{[Raja14b]}, which consists of images
from 28 semi-cooperative subjects (56 eyes)
captured using the rear camera of two different smartphones
(Apple iPhone 5S and Nokia Lumia 1020).
Images have been obtained in unconstrained conditions under mixed
illumination consisting of natural sunlight and artificial room
light. Each eye has 5 samples per smartphone, thus totalling
5$\times$56=280 images per device (560 in total).
Figure~\ref{fig:db-iris} shows some example images.
All images are resized via bicubic
interpolation to have the same iris radius
using MATLAB's \emph{imresize} function
(we choose as target
radius the average iris radius $R$=145 of the whole database,
given by available ground-truth).
Then, images are aligned by extracting a
square region of 319$\times$319 around the sclera center
(about 1.1$\times$$R$).
Two sample iris images after this procedure
can be seen in Figure~\ref{fig:images-example-iris},
right.

Aligned and normalized high-resolution images are then down-sampled
via bicubic interpolation to different sizes,
and then used as input low-resolution images
of the reconstruction methods.
The low-resolution images are then hallucinated
to the original input size.
Given an
input low-resolution image, we use all available images from the remaining eyes
(of both smartphones) to train the hallucination methods
(leave-one-out). Training images are mirrored in the
horizontal direction to duplicate the size of the training dataset,
thus having 55 eyes $\times$ 10 samples $\times$ 2 = 1100 images for
training.
Verification experiments are done separately for each device.
Each eye is considered as a different enrolled user.
As enrolment samples, we employ original high-resolution images,
whereas reconstructed images are employed as query data.
Genuine trials are done by pair-wise comparison
between all available images of the same eye, avoiding symmetric matches.
Impostor trials are done by comparing the first image of an eye
to the second image of the remaining eyes.
This procedure results in 56 $\times$ 10 = 560 genuine
and 56 $\times$ 55 = 3,018 impostor scores per device.

The results in Tables \ref{tbl:table_quality_iris} and \ref{tbl:table_recognition_iris}
show the performance of different iris super-resolution techniques mentioned in this chapter
in terms of both quality (PSNR and SSIM) and Equal Error Rate verification performance, respectively.
It can be observed that the two trained reconstruction methods evaluated outperform bicubic interpolation.
Its advantage is more evident at very high magnification factors, where the biggest differences
in quality metrics and verification performance occur.
An interesting observation is that the different evaluation metrics employed here do not
show the same tendency or relative difference as resolution changes.
For example,
the PSNR of bicubic interpolation is similar to that of the best trained method
up to a magnification factor of 8; but with bigger magnification factors,
the trained reconstruction methods achieve higher PSNR.
The SSIM, on the other hand, remains similar.
And despite the PSNR or SSIM being similar or not,
the verification performance of bicubic
is much worse than the trained methods,
regardless of the magnification factor employed.
This demonstrates again that image quality metrics are not
necessarily good predictors of the recognition performance \cite{[Alonso12a]}.

Regarding the two trained methods evaluated,
there is no clear winner.
Regarding the neighborhood size $k$ of LINE,
there are no conclusive results either.
The choice of $k$ does not seem to have a significant
impact on the performance.
Only with a magnification factor of 22,
there is a clear tendency for a bigger value of $k$.
It can be seen in Figure \ref{fig:images-example-iris}
that the images restored with a bigger $k$ are
more blurred (due to more patches being averaged),
but this seems to be positive
for the recognition performance nevertheless.
It is also worth noting that the verification performance
using trained reconstruction remains similar to the
baseline performance up to a magnification factor of 8
(which corresponds to an image size of only 29$\times$29).
This would allow to keep query images of very low size
without sacrificing performance,
with positive implications for example if
there are data storage or transmission restrictions.

\begin{table*}[!tb]
\centering
\caption{Summary of the Quality Analysis results using the PSNR and SSIM metrics on the VSSIRIS dataset.}
\label{tbl:table_quality_iris}
\begin{tabular}{|l||c|c||c|c||c|c||c|c||c|c|}
\multicolumn{1}{c}{} & \multicolumn{10}{c}{\bf{Magnification factor}} \\ \cline{2-11}
\multicolumn{1}{c}{} & \multicolumn{2}{c}{\bf{2}} & \multicolumn{2}{c}{\bf{4}} & \multicolumn{2}{c}{\bf{8}} & \multicolumn{2}{c}{\bf{16}} & \multicolumn{2}{c}{\bf{22}}\\
\cline{2-11}

\multicolumn{1}{c||}{\bf{SR Method}} & \bf{PSNR} & \bf{SSIM} & \bf{PSNR} & \bf{SSIM} & \bf{PSNR} & \bf{SSIM} & \bf{PSNR} & \bf{SSIM} & \bf{PSNR} & \bf{SSIM} \\
\hline

Bicubic 							 & 39.2 & \textbf{0.96} & \textbf{33.55} & \textbf{0.88} & 29.93 & \textbf{0.8} & 26.77 & \textbf{0.75} & 24.97 & 0.72  \\
Eigen-patches \cite{[Alonso15b]} & \textbf{39.4} & \textbf{0.96} & \textbf{33.53} & 0.87 & \textbf{30.19} & 0.79 & 27.37 & \textbf{0.75} & 26 & \textbf{0.73}  \\
LINE \cite{[Alonso17]} ($k = 75$) & 38.43 & 0.95 & 31.21 & 0.78 & 26.19 & 0.57 & 26.13 & 0.66 & 25.46 & 0.68  \\
LINE \cite{[Alonso17]} ($k = 150$) & 38.1 & 0.94 & 30.24 & 0.74 & 27.09 & 0.62 & 26.87 & 0.71 & 25.85 & 0.71  \\
LINE \cite{[Alonso17]} ($k = 300$) & 37.65 & 0.93 & 28.87 & 0.67 & 28.76 & 0.72 & 27.25 & 0.73 & 26.06 & 0.72  \\
LINE \cite{[Alonso17]} ($k = 600$) & 36.8 & 0.92 & 30.36 & 0.74 & 29.57 & 0.76 & 27.41 & 0.74 & \textbf{26.15} & \textbf{0.73}  \\
LINE \cite{[Alonso17]} ($k = 900$) & 35.92 & 0.9 & 31.59 & 0.79 & 29.82 & 0.77 & \textbf{27.45} & 0.74 & \textbf{26.17} & \textbf{0.73}  \\

\hline
\end{tabular}
\end{table*}

\begin{table*}[tb]
\centering
\caption{Summary of the EER recognition performance (in \%) using the Log-Gabor iris recognition algorithm \cite{[MasekThesis03]} on the VSSIRIS dataset.}
\label{tbl:table_recognition_iris}
\begin{tabular}{lc|c|c|c|c|c|c|c|c|c|c|c|}

\multicolumn{2}{c}{} & \multicolumn{5}{c}{\bf{iPhone}} & \multicolumn{1}{c}{} & \multicolumn{5}{c}{\bf{Nokia}} \\ \cline{3-7} \cline{9-13}
\multicolumn{2}{c}{} & \multicolumn{5}{c}{\bf{Magnification factor}} & \multicolumn{1}{c}{} & \multicolumn{5}{c}{\bf{Magnification factor}} \\ \cline{3-7} \cline{9-13}
\bf{SR Method} & & \bf{2} & \bf{4} & \bf{8} & \bf{16} & \bf{22} & & \bf{2} & \bf{4} & \bf{8} & \bf{16} & \bf{22} \\ \cline{3-7} \cline{9-13}
High-resolution & & \multicolumn{5}{c}{8.04} & \multicolumn{1}{|c|}{} & \multicolumn{5}{c}{7.5} \\ \cline{3-7} \cline{9-13}
Bicubic & & 14.47 & 13.88 & 14.79 & 16.14 & 18.47 & \multicolumn{1}{|c|}{} & 11.24 & 10.38 & 10.88 & 12.36 & 14.93 \\
Eigen-patches \cite{[Alonso15b]} & & 8.38 & 8.33 & \textbf{7.96} & \textbf{8.96} & 10.68    & \multicolumn{1}{|c|}{} & \textbf{7.61} & \textbf{7.09} & \textbf{7.3} & 9.55 & 10.54 \\
LINE \cite{[Alonso17]} ($k = 75$) & &  \textbf{7.94} & 8.28 & 8.56 & 9.98 & 13.55  & \multicolumn{1}{|c|}{} & 7.67 & 8 & 8.05 & \textbf{9.27} & 12.65 \\
LINE \cite{[Alonso17]} ($k = 150$) & & 8.17 & 8.73 & 8.12 & 9.59 & 12.55   & \multicolumn{1}{|c|}{} & 7.75 & 8.03 & 8.19 & 9.81 & 11.79 \\
LINE \cite{[Alonso17]} ($k = 300$) & & 8.17 & 8.52 & 8.88 & 9.57 & 12   & \multicolumn{1}{|c|}{} & 7.65 & 8.56 & 7.81 & 9.44 & 10.75 \\
LINE \cite{[Alonso17]} ($k = 600$) & & 8.03 & 8.77 & 8.54 & 9.59 & 11.53   & \multicolumn{1}{|c|}{} & 7.65 & 7.95 & 7.62 & 9.98 & \textbf{10.22} \\
LINE \cite{[Alonso17]} ($k = 900$) & & 8.03 & \textbf{8.11} & 8.33 & 9.61 & \textbf{10.59}   & \multicolumn{1}{|c|}{} & 7.75 & 7.71 & 7.64 & 9.37 & 10.7 \\  \cline{3-7} \cline{9-13}
\end{tabular}
\end{table*}

\section{Summary and Future Trends}

Face and iris biometrics are two well-explored modalities, with systems
yielding state-of-the-art performance in controlled scenarios.
However, the use of more relaxed acquisition environments,
like the one represented in selfie biometrics, is pushing
image-based biometrics towards the use of low-resolution
imagery. If not tackled properly, low resolution images
can pose significant problems in terms of
reduced performance. In this context,
super-resolution techniques can be used to enhance the
quality of low-resolution images to improve the recognition
performance of existing biometric systems.

Super-resolution is a core topic in computer vision,
with many techniques proposed to restore
low-resolution images \cite{Nasrollahi2014,[Thapa16]}.
However,
compared with the existing literature in generic super-resolution,
super-resolution in biometrics is a relatively recent topic \cite{[Nguyen18a]}.
This is because most approaches are general-scene, designed
to produce overall visual enhancement.
They try to improve
the quality of the image by minimizing objective measures,
such as the Peak Signal-to-Noise (PSNR), which does not
necessarily correlate with better recognition performance \cite{[Alonso13]}.
Images from a specific biometric modality have particular local and global structures
that can be exploited to achieve a more efficient up-sampling \cite{[Baker02]}.
For example, recovering local texture details is essential for face and ocular images
due to the prevalence of texture-based recognition in these modalities \cite{[Jain16]}.
This chapter has presented an overview of the image super-resolution topic,
with emphasis on the reconstruction of face and iris images in the visible spectrum,
which are the two prevalent modalities in selfie biometrics.
We investigate several existing techniques,
and evaluate their application to reconstruct face and iris images.

Despite promising performance of super-resolution methods
for facial or ocular images under well-controlled conditions,
they degenerate when encountering images from uncontrolled environments,
as for example non-frontal views, expression or lightning changes \cite{[Wang14]}.
Future trends in biometrics super-resolution therefore
relate to designing effective approaches to cope with these variations.
For example, one limitation of existing studies is that low-resolution images are
simulated by down-sampling high-resolution images
due to the lack of databases with
low-resolution and corresponding high-resolution reference images.
As a result, variations in pose, illumination or expression
are not yet fully considered,
neither the associated artifacts introduced (e.g. compression, noise or blur).
In addition, prior to down-sampling,
images are aligned by manual annotation of landmarks (eyes, nose, etc.)
followed by affine transformation.
All super-resolution schemes employed in the biometric literature
are heavily affected by imprecise image alignment, even by small amounts;
however, real low-resolution images usually have blurring,
and so many ambiguities exist for landmark localization or segmentation,
thus making a necessity the use of reconstruction schemes
capable of working under imprecise alignment.

\section*{Acknowledgment}

Authors F. A.-F. and J. B thank the Swedish Research Council (VR), the Sweden's innovation agency (VINNOVA),
and the Swedish Knowledge Foundation (CAISR program and SIDUS-AIR project).
Author J. F. thanks Accenture and the project CogniMetrics (TEC2015-70627-R) from MINECO/FEDER,

\bibliographystyle{spmpsci}

\begin{thebibliography}{10}
\providecommand{\url}[1]{{#1}}
\providecommand{\urlprefix}{URL }
\expandafter\ifx\csname urlstyle\endcsname\relax
  \providecommand{\doi}[1]{DOI~\discretionary{}{}{}#1}\else
  \providecommand{\doi}{DOI~\discretionary{}{}{}\begingroup
  \urlstyle{rm}\Url}\fi

\bibitem{Ahonen2006}
Ahonen, T., Hadid, A., Pietikainen, M.: Face description with local binary
  patterns: Application to face recognition.
\newblock IEEE Transactions on Pattern Analysis and Machine Intelligence
  \textbf{28}(12), 2037--2041 (2006).
\newblock \doi{10.1109/TPAMI.2006.244}

\bibitem{[Aljadaany15]}
Aljadaany, R., Luu, K., Venugopalan, S., Savvides, M.: Iris super-resolution
  via nonparametric over-complete dictionary learning.
\newblock Proc IEEE Intl Conf on Image Processing, ICIP pp. 3856--3860 (2015).
\newblock \doi{10.1109/ICIP.2015.7351527}

\bibitem{[Alonso13]}
{Alonso-Fernandez}, F., Bigun, J.: Quality factors affecting iris segmentation
  and matching.
\newblock In: Proc Intl Conf on Biometrics, ICB, pp. 1--6 (2013).
\newblock \doi{10.1109/ICB.2013.6613016}

\bibitem{[Alonso15b]}
{Alonso-Fernandez}, F., Farrugia, R.A., Bigun, J.: Eigen-patch iris
  super-resolution for iris recognition improvement.
\newblock Proc European Signal Processing Conference, EUSIPCO  (2015)

\bibitem{[Alonso17]}
Alonso-Fernandez, F., Farrugia, R.A., Bigun, J.: Iris super-resolution using
  iterative neighbor embedding.
\newblock Proc IEEE Conference on Computer Vision and Pattern Recognition
  Workshops, CVPRW pp. 655--663 (2017).
\newblock \doi{10.1109/CVPRW.2017.94}

\bibitem{[Alonso12a]}
{Alonso-Fernandez}, F., Fierrez, J., {Ortega-Garcia}, J.: Quality measures in
  biometric systems.
\newblock IEEE Security and Privacy \textbf{10}(6), 52--62 (2012)

\bibitem{Baker2000}
Baker, S., Kanade, T.: Hallucinating faces.
\newblock In: Proceedings Fourth IEEE International Conference on Automatic
  Face and Gesture Recognition (Cat. No. PR00580), pp. 83--88 (2000).
\newblock \doi{10.1109/AFGR.2000.840616}

\bibitem{[Baker02]}
Baker, S., Kanade, T.: Limits on super-resolution and how to break them.
\newblock IEEE Transactions on Pattern Analysis and Machine Intelligence
  \textbf{24}(9), 1167--1183 (2002)

\bibitem{[Barnard06]}
Barnard, R., Pauca, V.P., Torgersen, T.C., Plemmons, R.J., Prasad, S., van~der
  Gracht, J., Nagy, J., Chung, J., Behrmann, G., Mathews, S., Mirotznik, M.:
  High-resolution iris image reconstruction from low-resolution imagery (2006).
\newblock \doi{10.1117/12.681930}.
\newblock \urlprefix\url{https://doi.org/10.1117/12.681930}

\bibitem{Cao2017}
Cao, Q., Lin, L., Shi, Y., Liang, X., Li, G.: Attention-aware face
  hallucination via deep reinforcement learning.
\newblock In: 2017 IEEE Conference on Computer Vision and Pattern Recognition
  (CVPR), pp. 1656--1664 (2017).
\newblock \doi{10.1109/CVPR.2017.180}

\bibitem{Carrato1996}
Carrato, S., Ramponi, G., Marsi, S.: A simple edge-sensitive image
  interpolation filter.
\newblock In: Proceedings of 3rd IEEE International Conference on Image
  Processing, vol.~3, pp. 711--714 vol.3 (1996)

\bibitem{Chakrabarti2007}
Chakrabarti, A., Rajagopalan, A.N., Chellappa, R.: Super-resolution of face
  images using kernel pca-based prior.
\newblock IEEE Transactions on Multimedia \textbf{9}(4), 888--892 (2007).
\newblock \doi{10.1109/TMM.2007.893346}

\bibitem{Chang2004}
Chang, H., Yeung, D.Y., Xiong, Y.: Super-resolution through neighbor embedding.
\newblock In: Proceedings of the 2004 IEEE Computer Society Conference on
  Computer Vision and Pattern Recognition, 2004. CVPR 2004., vol.~1, pp. I--I
  (2004)

\bibitem{Cheeseman1996}
Cheeseman, P., Kanefsky, B., Kraft, R., Stutz, J., Hanson, R.: Super-Resolved
  Surface Reconstruction from Multiple Images, pp. 293--308.
\newblock Springer Netherlands, Dordrecht (1996)

\bibitem{[Chen14]}
Chen, H.Y., Chien, S.Y.: Eigen-patch: Position-patch based face hallucination
  using eigen transformation.
\newblock In: Proc. IEEE Intl. Conf. on Multimedia and Expo, ICME, pp. 1--6
  (2014)

\bibitem{[Deshpande17]}
Deshpande, A., Patavardhan, P.: Multi-frame super-resolution for long range
  captured iris polar image.
\newblock IET Biometrics \textbf{6}(2), 108--116 (2017).
\newblock \doi{10.1049/iet-bmt.2016.0076}

\bibitem{[Deshpande17a]}
Deshpande, A., Patavardhan, P.P.: Super resolution and recognition of long
  range captured multi-frame iris images.
\newblock IET Biometrics \textbf{6}(5), 360--368 (2017).
\newblock \doi{10.1049/iet-bmt.2016.0075}

\bibitem{Dong2015}
Dong, C., Loy, C.C., He, K., Tang, X.: Image super-resolution using deep
  convolutional networks.
\newblock IEEE Transactions on Pattern Analysis and Machine Intelligence
  \textbf{38}(2), 295--307 (2016).
\newblock \doi{10.1109/TPAMI.2015.2439281}

\bibitem{Dong2013}
Dong, W., Zhang, L., Shi, G., Li, X.: Nonlocally centralized sparse
  representation for image restoration.
\newblock IEEE Transactions on Image Processing \textbf{22}(4), 1620--1630
  (2013).
\newblock \doi{10.1109/TIP.2012.2235847}

\bibitem{[Fahmy07]}
Fahmy, G.: Super-resolution construction of iris images from a visual low
  resolution face video.
\newblock Proc National Radio Science Conference, NRSC  (2007)

\bibitem{Farrugia2016}
Farrugia, R.A., Guillemot, C.: Robust face hallucination using
  quantization-adaptive dictionaries.
\newblock In: 2016 IEEE International Conference on Image Processing (ICIP),
  pp. 414--418 (2016).
\newblock \doi{10.1109/ICIP.2016.7532390}

\bibitem{[Farrugia17]}
Farrugia, R.A., Guillemot, C.: Face hallucination using linear models of
  coupled sparse support.
\newblock IEEE Transactions on Image Processing \textbf{26}(9), 4562--4577
  (2017).
\newblock \doi{10.1109/TIP.2017.2717181}

\bibitem{Farrugia2017}
Farrugia, R.A., Guillemot, C.: Face hallucination using linear models of
  coupled sparse support.
\newblock IEEE Transactions on Image Processing \textbf{26}(9), 4562--4577
  (2017).
\newblock \doi{10.1109/TIP.2017.2717181}

\bibitem{Farsiu2003}
Farsiu, S., Robinson, D., Elad, M., Milanfar, P.: Fast and robust
  super-resolution.
\newblock In: Proceedings 2003 International Conference on Image Processing
  (Cat. No.03CH37429), vol.~2, pp. II--291--4 vol.3 (2003).
\newblock \doi{10.1109/ICIP.2003.1246674}

\bibitem{Farsiu2003b}
Farsiu, S., Robinson, D., Elad, M., Milanfar, P.: Robust shift and add approach
  to superresolution (2003).
\newblock \doi{10.1117/12.507194}

\bibitem{Farsiu2004}
Farsiu, S., Robinson, M.D., Elad, M., Milanfar, P.: Fast and robust multiframe
  super resolution.
\newblock IEEE Transactions on Image Processing \textbf{13}(10), 1327--1344
  (2004).
\newblock \doi{10.1109/TIP.2004.834669}

\bibitem{[Fierrez18]}
Fierrez, J., Morales, A., Vera-Rodriguez, R., Camacho, D.: Multiple classifiers
  in biometrics. part 1: Fundamentals and review.
\newblock Information Fusion \textbf{44}, 57--64 (2018).
\newblock \doi{https://doi.org/10.1016/j.inffus.2017.12.003}

\bibitem{[Fierrez18a]}
Fierrez, J., Morales, A., Vera-Rodriguez, R., Camacho, D.: Multiple classifiers
  in biometrics. part 2: Trends and challenges.
\newblock Information Fusion \textbf{44}, 103--112 (2018).
\newblock \doi{https://doi.org/10.1016/j.inffus.2017.12.005}

\bibitem{Freeman2002}
Freeman, W.T., Jones, T.R., Pasztor, E.C.: Example-based super-resolution.
\newblock IEEE Computer Graphics and Applications \textbf{22}(2), 56--65
  (2002).
\newblock \doi{10.1109/38.988747}

\bibitem{[Galbally14]}
Galbally, J., Marcel, S., Fierrez, J.: Image quality assessment for fake
  biometric detection: Application to iris, fingerprint, and face recognition.
\newblock IEEE Transactions on Image Processing \textbf{23}(2), 710--724 (2014)

\bibitem{Glasner2009}
Glasner, D., Bagon, S., Irani, M.: Super-resolution from a single image.
\newblock In: The IEEE International Conference on Computer Vision (ICCV)
  (2009)

\bibitem{Gonzalez2006}
Gonzalez, R.C., Woods, R.E.: Digital Image Processing (3rd Edition).
\newblock Prentice-Hall, Inc., Upper Saddle River, NJ, USA (2006)

\bibitem{Hardie1997}
Hardie, R.C., Barnard, K.J., Armstrong, E.E.: Joint map registration and
  high-resolution image estimation using a sequence of undersampled images.
\newblock IEEE Transactions on Image Processing \textbf{6}(12), 1621--1633
  (1997).
\newblock \doi{10.1109/83.650116}

\bibitem{[Hollingsworth09a]}
Hollingsworth, K., Peters, T., Bowyer, K., Flynn, P.: Iris recognition using
  signal-level fusion of frames from video.
\newblock IEEE Transactions on Information Forensics and Security
  \textbf{4}(4), 837--848 (2009)

\bibitem{[Hsieh16]}
Hsieh, S.H., Li, Y.H., Tien, C.H., Chang, C.C.: Extending the capture volume of
  an iris recognition system using wavefront coding and super-resolution.
\newblock IEEE Transactions on Cybernetics \textbf{46}(12), 3342--3350 (2016).
\newblock \doi{10.1109/TCYB.2015.2504388}

\bibitem{Hu2011}
Hu, Y., Lam, K.M., Qiu, G., Shen, T.: From local pixel structure to global
  image super-resolution: A new face hallucination framework.
\newblock IEEE Transactions on Image Processing \textbf{20}(2), 433--445
  (2011).
\newblock \doi{10.1109/TIP.2010.2063437}

\bibitem{Huang2010}
Huang, H., He, H., Fan, X., Zhang, J.: Super-resolution of human face image
  using canonical correlation analysis.
\newblock Pattern Recogn. \textbf{43}(7), 2532--2543 (2010).
\newblock \doi{10.1016/j.patcog.2010.02.007}.
\newblock \urlprefix\url{http://dx.doi.org/10.1016/j.patcog.2010.02.007}

\bibitem{[Huang03]}
Huang, J., Ma, L., Tan, T., Wang, Y.: Learning based resolution enhancement of
  iris images.
\newblock Proc. BMVC  (2003)

\bibitem{Irani1990}
Irani, M., Peleg, S.: Super resolution from image sequences.
\newblock In: [1990] Proceedings. 10th International Conference on Pattern
  Recognition, vol.~ii, pp. 115--120 vol.2 (1990).
\newblock \doi{10.1109/ICPR.1990.119340}

\bibitem{Irani1993}
Irani, M., Peleg, S.: Motion analysis for image enhancement: Resolution,
  occlusion, and transparency.
\newblock Journal of Visual Communication and Image Representation
  \textbf{4}(4), 324 -- 335 (1993).
\newblock \doi{https://doi.org/10.1006/jvci.1993.1030}.
\newblock
  \urlprefix\url{http://www.sciencedirect.com/science/article/pii/S1047320383710308}

\bibitem{[Jain16]}
Jain, A., Nandakumar, K., Ross, A.: 50 years of biometric research:
  Accomplishments, challenges, and opportunities.
\newblock Pattern Recognition Letters \textbf{79}, 80--105 (2016)

\bibitem{[Jain11a]}
Jain, A.K., Kumar, A.: Second Generation Biometrics, chap. {B}iometrics of
  {N}ext {G}eneration: {A}n {O}verview.
\newblock Springer (2011)

\bibitem{Jiang2014}
Jiang, J., Hu, R., Wang, Z., Han, Z.: Face super-resolution via multilayer
  locality-constrained iterative neighbor embedding and intermediate dictionary
  learning.
\newblock IEEE Transactions on Image Processing \textbf{23}(10), 4220--4231
  (2014).
\newblock \doi{10.1109/TIP.2014.2347201}

\bibitem{[Jillela11]}
Jillela, R., Ross, A., Flynn, P.: Information fusion in low-resolution iris
  videos using principal components transform.
\newblock Proc. IEEE Workshop on Applications of Computer Vision, WACV pp.
  262--269 (2011).
\newblock \doi{10.1109/WACV.2011.5711512}

\bibitem{Jung2011}
Jung, C., Jiao, L., Liu, B., Gong, M.: Position-patch based face hallucination
  using convex optimization.
\newblock IEEE Signal Processing Letters \textbf{18}(6), 367--370 (2011).
\newblock \doi{10.1109/LSP.2011.2140370}

\bibitem{Kim2016}
Kim, J., Kwon~Lee, J., Mu~Lee, K.: Accurate image super-resolution using very
  deep convolutional networks.
\newblock In: The IEEE Conference on Computer Vision and Pattern Recognition
  (CVPR) (2016)

\bibitem{Li2009}
Li, B., Chang, H., Shan, S., Chen, X.: Aligning coupled manifolds for face
  hallucination.
\newblock IEEE Signal Processing Letters \textbf{16}(11), 957--960 (2009).
\newblock \doi{10.1109/LSP.2009.2027657}

\bibitem{Li2014}
Li, Y., Cai, C., Qiu, G., Lam, K.M.: Face hallucination based on sparse
  local-pixel structure.
\newblock Pattern Recognition \textbf{47}(3), 1261 -- 1270 (2014).
\newblock \doi{https://doi.org/10.1016/j.patcog.2013.09.012}.
\newblock
  \urlprefix\url{http://www.sciencedirect.com/science/article/pii/S0031320313003841}.
\newblock Handwriting Recognition and other PR Applications

\bibitem{Lim2017}
Lim, B., Son, S., Kim, H., Nah, S., Lee, K.M.: Enhanced deep residual networks
  for single image super-resolution.
\newblock In: 2017 IEEE Conference on Computer Vision and Pattern Recognition
  Workshops (CVPRW), pp. 1132--1140 (2017).
\newblock \doi{10.1109/CVPRW.2017.151}

\bibitem{Lin2004}
Lin, Z., Shum, H.Y.: Fundamental limits of reconstruction-based superresolution
  algorithms under local translation.
\newblock IEEE Trans. Pattern Anal. Mach. Intell. \textbf{26}(1), 83--97 (2004)

\bibitem{[Liu13]}
Liu, J., Sun, Z., Tan, T.: Code-level information fusion of low-resolution iris
  image sequences for personal identification at a distance.
\newblock Proc Intl Conf on Biometrics: Theory, Applications and Systems, BTAS
  pp. 1--6 (2013).
\newblock \doi{10.1109/BTAS.2013.6712692}

\bibitem{Ma2009}
Ma, X., Zhang, J., Qi, C.: Position-based face hallucination method.
\newblock In: 2009 IEEE International Conference on Multimedia and Expo, pp.
  290--293 (2009).
\newblock \doi{10.1109/ICME.2009.5202492}

\bibitem{[MasekThesis03]}
Masek, L.: Recognition of human iris patterns for biometric identification.
\newblock Master's thesis, School of Computer Science and Software Engineering,
  University of Western Australia (2003)

\bibitem{Nasir2011}
Nasir, H., Stankovic, V., Marshall, S.: Singular value decomposition based
  fusion for super-resolution image reconstruction.
\newblock In: 2011 IEEE International Conference on Signal and Image Processing
  Applications (ICSIPA), pp. 393--398 (2011).
\newblock \doi{10.1109/ICSIPA.2011.6144138}

\bibitem{Nasrollahi2014}
Nasrollahi, K., Moeslund, T.B.: Super-resolution: a comprehensive survey.
\newblock Machine Vision and Applications \textbf{25}(6), 1423--1468 (2014).
\newblock \doi{10.1007/s00138-014-0623-4}.
\newblock \urlprefix\url{https://doi.org/10.1007/s00138-014-0623-4}

\bibitem{[Nguyen10a]}
Nguyen, K., Fookes, C., Sridharan, S.: Robust mean super-resolution for less
  cooperative nir iris recognition at a distance and on the move.
\newblock Proc Symposium on Information and Communication Technology, SoICT pp.
  122--127 (2010).
\newblock \doi{10.1145/1852611.1852635}.
\newblock \urlprefix\url{http://doi.acm.org/10.1145/1852611.1852635}

\bibitem{[Nguyen10]}
Nguyen, K., Fookes, C., Sridharan, S., Denman, S.: Focus-score weighted
  super-resolution for uncooperative iris recognition at a distance and on the
  move.
\newblock Proc. 25th International Conference of Image and Vision Computing New
  Zealand, IVCNZ pp. 1--8 (2010).
\newblock \doi{10.1109/IVCNZ.2010.6148792}

\bibitem{[Nguyen11a]}
Nguyen, K., Fookes, C., Sridharan, S., Denman, S.: Feature-domain
  super-resolution for iris recognition.
\newblock Proc IEEE International Conference on Image Processing, ICIP pp.
  3197--3200 (2011).
\newblock \doi{10.1109/ICIP.2011.6116348}

\bibitem{[Nguyen11]}
Nguyen, K., Fookes, C., Sridharan, S., Denman, S.: Quality-driven
  super-resolution for less constrained iris recognition at a distance and on
  the move.
\newblock IEEE Transactions on Information Forensics and Security
  \textbf{6}(4), 1248--1258 (2011)

\bibitem{[Nguyen13]}
Nguyen, K., Fookes, C., Sridharan, S., Denman, S.: Feature-domain
  super-resolution for iris recognition.
\newblock Computer Vision and Image Understanding \textbf{117}(10), 1526 --
  1535 (2013).
\newblock \doi{https://doi.org/10.1016/j.cviu.2013.06.010}.
\newblock
  \urlprefix\url{http://www.sciencedirect.com/science/article/pii/S1077314213001306}

\bibitem{[Nguyen18a]}
Nguyen, K., Fookes, C., Sridharan, S., Tistarelli, M., Nixon, M.:
  Super-resolution for biometrics: A comprehensive survey.
\newblock Pattern Recognition \textbf{78}, 23 -- 42 (2018).
\newblock \doi{https://doi.org/10.1016/j.patcog.2018.01.002}.
\newblock
  \urlprefix\url{http://www.sciencedirect.com/science/article/pii/S0031320318300049}

\bibitem{[Nguyen12]}
Nguyen, K., Sridharan, S., Denman, S., Fookes, C.: Feature-domain
  super-resolution framework for gabor-based face and iris recognition.
\newblock Proc IEEE Conf on Computer Vision and Pattern Recognition, CVPR pp.
  2642--2649 (2012)

\bibitem{[Othman15]}
Othman, N., Dorizzi, B.: Impact of quality-based fusion techniques for
  video-based iris recognition at a distance.
\newblock IEEE Transactions on Information Forensics and Security
  \textbf{10}(8), 1590--1602 (2015).
\newblock \doi{10.1109/TIFS.2015.2421314}

\bibitem{Park2008}
Park, J.S., Lee, S.W.: An example-based face hallucination method for
  single-frame, low-resolution facial images.
\newblock IEEE Transactions on Image Processing \textbf{17}(10), 1806--1816
  (2008).
\newblock \doi{10.1109/TIP.2008.2001394}

\bibitem{[Parkhi15]}
Parkhi, O.M., Vedaldi, A., Zisserman, A.: Deep face recognition.
\newblock Proc British Machine Vision Conference, BMVC  (2015)

\bibitem{Pham2006}
Pham, T.Q., van Vliet, L.J., Schutte, K.: Robust fusion of irregularly sampled
  data using adaptive normalized convolution.
\newblock EURASIP Journal on Advances in Signal Processing \textbf{2006}(1),
  083,268 (2006).
\newblock \doi{10.1155/ASP/2006/83268}.
\newblock \urlprefix\url{https://doi.org/10.1155/ASP/2006/83268}

\bibitem{[Raja14b]}
Raja, K.B., Raghavendra, R., Vemuri, V.K., Busch, C.: Smartphone based visible
  iris recognition using deep sparse filtering.
\newblock Pattern Recognition Letters \textbf{57}, 33--42 (2015)

\bibitem{[Ribeiro17]}
Ribeiro, E., Uhl, A., Alonso-Fernandez, F., Farrugia, R.A.: Exploring deep
  learning image super-resolution for iris recognition.
\newblock Proc 25th European Signal Processing Conference, EUSIPCO pp.
  2176--2180 (2017).
\newblock \doi{10.23919/EUSIPCO.2017.8081595}

\bibitem{Schultz1996}
Schultz, R.R., Stevenson, R.L.: Extraction of high-resolution frames from video
  sequences.
\newblock IEEE Transactions on Image Processing \textbf{5}(6), 996--1011
  (1996).
\newblock \doi{10.1109/83.503915}

\bibitem{[Shin09]}
Shin, K.Y., Park, K.R., Kang, B.J., Park, S.J.: Super-resolution method based
  on multiple multi-layer perceptrons for iris recognition.
\newblock In: Intl Conf Ubiquitous Information Technologies Applications, ICUT,
  pp. 1--5 (2009)

\bibitem{Simonyan2008}
Simonyan, K., Grishin, S., Vatolin, D., Popov, D.: Fast video super-resolution
  via classification.
\newblock In: 2008 15th IEEE International Conference on Image Processing, pp.
  349--352 (2008).
\newblock \doi{10.1109/ICIP.2008.4711763}

\bibitem{Su2004}
Su, D., Willis, P.: {Image Interpolation by Pixel-Level Data-Dependent
  Triangulation}.
\newblock Computer Graphics Forum  (2004).
\newblock \doi{10.1111/j.1467-8659.2004.00752.x}

\bibitem{Su2005}
Su, K., Tian, Q., Xue, Q., Sebe, N., Ma, J.: Neighborhood issue in single-frame
  image super-resolution.
\newblock In: 2005 IEEE International Conference on Multimedia and Expo, pp. 4
  pp.-- (2005).
\newblock \doi{10.1109/ICME.2005.1521623}

\bibitem{[Thapa16]}
Thapa, D., Raahemifar, K., Bobier, W.R., Lakshminarayanan, V.: A performance
  comparison among different super-resolution techniques.
\newblock Computers \& Electrical Engineering \textbf{54}, 313 -- 329 (2016).
\newblock \doi{https://doi.org/10.1016/j.compeleceng.2015.09.011}.
\newblock
  \urlprefix\url{http://www.sciencedirect.com/science/article/pii/S0045790615003183}

\bibitem{Thevenaz2000}
Th{\'e}venaz, P., Blu, T., Unser, M.: Handbook of medical imaging.
\newblock chap. Image Interpolation and Resampling, pp. 393--420. Academic
  Press, Inc., Orlando, FL, USA (2000).
\newblock \urlprefix\url{http://dl.acm.org/citation.cfm?id=374166.374424}

\bibitem{[Turk91]}
Turk, M., Pentland, A.: Eigenfaces for recognition.
\newblock Journal of Cognitive Neuroscience \textbf{3}(1), 71--86 (1991)

\bibitem{[Wang14]}
Wang, N., Tao, D., Gao, X., Li, X., Li, J.: A comprehensive survey to face
  hallucination.
\newblock Intl Journal of Computer Vision \textbf{106}(1), 9--30 (2014)

\bibitem{Wang2005}
Wang, X., Tang, X.: Hallucinating face by eigentransformation.
\newblock IEEE Transactions on Systems, Man, and Cybernetics, Part C
  (Applications and Reviews) \textbf{35}(3), 425--434 (2005).
\newblock \doi{10.1109/TSMCC.2005.848171}

\bibitem{[Wang04]}
Wang, Z., Bovik, A.C., Sheikh, H.R., Simoncelli, E.P.: Image quality
  assessment: from error visibility to structural similarity.
\newblock IEEE Transactions on Image Processing \textbf{13}(4), 600--612
  (2004).
\newblock \doi{10.1109/TIP.2003.819861}

\bibitem{Yang2008}
Yang, J., Wright, J., Huang, T., Ma, Y.: Image super-resolution as sparse
  representation of raw image patches.
\newblock In: 2008 IEEE Conference on Computer Vision and Pattern Recognition,
  pp. 1--8 (2008).
\newblock \doi{10.1109/CVPR.2008.4587647}

\bibitem{Yang2010}
Yang, J., Wright, J., Huang, T.S., Ma, Y.: Image super-resolution via sparse
  representation.
\newblock IEEE Transactions on Image Processing \textbf{19}(11), 2861--2873
  (2010).
\newblock \doi{10.1109/TIP.2010.2050625}

\bibitem{[Zhang16a]}
Zhang, Q., Li, H., He, Z., Sun, Z.: Image super-resolution for mobile iris
  recognition.
\newblock Proc 11th Chinese Conference on Biometric Recognition, CCBR pp.
  399--406 (2016)

\bibitem{Zomet2000}
Zomet, A., Peleg, S.: Efficient super-resolution and applications to mosaics.
\newblock In: Proceedings 15th International Conference on Pattern Recognition.
  ICPR-2000, vol.~1, pp. 579--583 vol.1 (2000).
\newblock \doi{10.1109/ICPR.2000.905404}

\end{thebibliography}

\end{document}